\documentclass{article}

\usepackage{amsmath,amsfonts,bm}

\newtheorem{theorem}{Theorem}[section]
\newtheorem{proposition}[theorem]{Proposition}
\newtheorem{lemma}[theorem]{Lemma}

\newtheorem{definition}[theorem]{Definition}

\def\eqref#1{equation~\ref{#1}}

\def\1{\bm{1}}

\DeclareMathAlphabet{\mathsfit}{\encodingdefault}{\sfdefault}{m}{sl}
\SetMathAlphabet{\mathsfit}{bold}{\encodingdefault}{\sfdefault}{bx}{n}

\newcommand{\R}{\mathbb{R}}

\PassOptionsToPackage{round}{natbib}

\usepackage[final]{neurips_2025}

\usepackage[utf8]{inputenc} 
\usepackage[T1]{fontenc}    
\usepackage{hyperref}       
\usepackage{url}            
\usepackage{booktabs}       
\usepackage{amsfonts}       
\usepackage{nicefrac}       
\usepackage{microtype}      
\usepackage{xcolor}         

\usepackage{algorithm}
\usepackage{algorithmic}
\usepackage{wrapfig}
\usepackage{tikz}
\usepackage{pgfplots}
    \pgfplotsset{
        discard if not/.style 2 args={
            filter discard warning=false,
            x filter/.append code={
                \edef\tempa{\thisrow{#1}}
                \edef\tempb{#2}
                \ifx\tempa\tempb
                \else
                    
                \fi
            },
        },
    }
    \pgfplotsset{compat=1.18}
\usepgfplotslibrary{statistics}
\usepackage{subfigure}

\title{Rao-Blackwellised Reparameterisation Gradients}

\author{
  Kevin H. Lam\\
  Department of Statistics\\
  University of Oxford\\
  \And
  Thang D. Bui\\
  School of Computing \\
  Australian National University \\
  \AND
  George Deligiannidis\\
  Department of Statistics\\
  University of Oxford\\
  \And
  Yee Whye Teh\\
  Department of Statistics\\
  University of Oxford\\
  \And
  \texttt{\{lam,deligian,y.w.teh\}@stats.ox.ac.uk, thang.bui@anu.edu.au}
}

\begin{document}

\maketitle

\begin{abstract}
Latent Gaussian variables have been popularised in probabilistic machine learning. In turn, gradient estimators are the machinery that facilitates gradient-based optimisation for models with latent Gaussian variables. The reparameterisation trick is often used as the default estimator as it is simple to implement and yields low-variance gradients for variational inference. In this work, we propose the R2-G2 estimator as the Rao-Blackwellisation of the reparameterisation gradient estimator. Interestingly, we show that the local reparameterisation gradient estimator for Bayesian MLPs is an instance of the R2-G2 estimator and Rao-Blackwellisation. This lets us extend benefits of Rao-Blackwellised gradients to a suite of probabilistic models. We show that initial training with R2-G2 consistently yields better performance in models with multiple applications of the reparameterisation trick.
\end{abstract}

\section{Introduction}
\label{section:introduction}
    Latent random variables are ubiquitous in probabilistic machine learning (ML) as they enable us to embed prior assumptions into models, such as uncertainty in model parameters and low-dimensional latent representations of observed data structures. Thus, latent random variables appear in a range of modelling tasks, including variational inference, generative modelling, and density estimation. A wide variety of modern models with latent random variables, particularly those that parameterise a neural network (NN), are trained with gradient-based optimisation. This presents a non-trivial problem on how analytical derivatives can incorporate stochasticity, and has led to substantial interest in \emph{gradient estimators}. These estimators enable gradient-based optimisation by computing approximations of gradients that are compatible with automatic differentiation software. For comprehensive reviews of gradient estimators with latent continuous and discrete random variables, we refer the reader to \cite{mohamed2020survey} and \cite{huijben2023survey} respectively.

We revisit the class of gradient estimators for latent Gaussian variables \citep{kingma2014vae,rezende2014stochastic,titsias2014doubly}, as Gaussian variables often serve as the default distribution for the noise or prior in probabilistic machine learning tasks. These estimators, known as pathwise estimators or the reparameterisation trick, have been popularised as they produce low-variance gradients for variational inference. In particular, it has been shown in \cite{xu2019variance} that the reparameterisation gradient estimator has lower variance than a Rao-Blackwellised REINFORCE estimator \citep{williams1992reinforce} for variational inference. These estimators are also attractive as they are \emph{single-sample} estimators since training models with them only requires sampling operations be done once. In turn, this also limits costly function evaluations needed for training as models grow larger in the era of deep learning.

In this work, we present the R2-G2 estimator as an extension of the reparameterisation gradient estimator formed by conditioning on pre-activations in backpropagation. Notably, we show that the local reparameterisation gradient estimator proposed in \cite{kingma2015local} is an instance of the R2-G2 estimator for linear layers in a Bayesian MLP with independent weights and establish that it is equivalent to a Rao-Blackwellised reparameterisation gradient estimator.

Our main contributions are as follows:
\begin{itemize}
    \item We present R2-G2, as a novel, general-purpose and single-sample gradient estimator for latent Gaussian variables as the Rao-Blackwellised reparameterisation gradient estimator;

    \item We show that the local reparameterisation gradient estimator is an instance of the R2-G2 estimator and enjoys variance reduction in gradients from Rao-Blackwellisation;
    
    \item We empirically show that initial training with the R2-G2 estimator consistently yields higher likelihoods on Bayesian Neural Networks and higher ELBOs on hierarchical Variational Autoencoders than the reparameterisation gradient estimator.
\end{itemize}

\section{Problem setting}
\label{section:problem_setting}
    Let $\mathbf{v} \in \R^n$ be a vector of continuous random variables where we have $\mathbf{v}\sim q_{\pmb{\theta}}$ for some probability distribution $q$ parameterised by a vector $\pmb{\theta}$. Suppose we are provided a continuously differentiable loss function $\ell:\R^n \to \R$ that depends on the random variables $\mathbf{v}$. To enable gradient-based training of a parametric model $f$, our goal is to compute the gradient of the expected loss
\begin{align}
\label{equation:general_gradient}
     \nabla_{\pmb{\theta}} \mathbb{E}_{q_{\pmb{\theta}}}[\ell_{\mathcal{D}, f}(\mathbf{v})]
\end{align}
where $\mathcal{D}$ is a dataset and $f$ is a parametric function optimised by the loss such as a neural network (NN), with both affecting the evaluation of $\ell$ (i.e. $\ell_{\mathcal{D}, f}$). We note that $\mathcal{D}$ is described generally here to encompass both supervised and unsupervised problems. For ease of notation in the remainder of this work, we will shorten our notation of the loss from $\ell_{\mathcal{D}, f}(\mathbf{v}, \pmb{\theta})$ to $\ell$ where the context is clear with its dependence on $\mathbf{v}$, $\pmb{\theta}$, $\mathcal{D}$, and $f$ implied. Popular choices of $\ell$ aim to maximise the log-likelihood or the evidence lower bound (ELBO) for the dataset $\mathcal{D}$. The former involves setting $\ell$ as the negative log-likelihood $\ell_{NLL}$. The latter sets $\ell$ as
\begin{align*}
     \ell_{ELBO}(\mathbf{v}) &=\ell_{NLL}(\mathcal{D}|\mathbf{v}) + \log\left(\frac{q_{\pmb{\theta}}(\mathbf{v}|\mathcal{D})}{p(\mathbf{v})}\right)
\end{align*}
where $p$ is a prior distribution of $\mathbf{v}$ and setting $q_{\pmb{\theta}}$ as the posterior of $\mathbf{v}$ or its approximation, making Equation \ref{equation:general_gradient} equivalent to maximising the ELBO. Moreover, it is equivalent to variational Bayesian inference when the form of $q_{\pmb{\theta}}$ is restricted as it becomes an approximation of the posterior.

In practice, we use a Monte Carlo approximation $\mathbb{E}_{q_{\pmb{\theta}}}[\ell(\mathbf{v}, \pmb{\theta})] \approx \frac{1}{M} \sum_{i=1}^{M} \ell(\mathbf{v}^{(i)}, \pmb{\theta})$ to write
\begin{align*}
     \nabla_{\pmb{\theta}} \mathbb{E}_{q_{\pmb{\theta}}}[\ell(\mathbf{v})] \approx \frac{1}{M}\sum_{i=1}^{M} \nabla_{\pmb{\theta}} \ell(\mathbf{v}^{(i)})
\end{align*}

where $\mathbf{v}^{(1)},\dots,\mathbf{v}^{(M)} \sim q_{\pmb{\theta}}$. However, this would require $M$ evaluations of the loss $\ell$ which is not desirable when the underlying model $f$ is large or costly to evaluate. In this work, we focus on the setting where $\mathbf{v}$ are \emph{independent} Gaussian random variables with the aim to derive a \emph{single-sample} gradient estimator (i.e. $M=1$) that is unbiased and enjoys reduced variance.

\section{Related work}
\label{section:related_work}
    In this section, we revisit the single-sample gradient estimators compatible with Gaussian random variables, namely REINFORCE, reparameterisation trick and local reparameterisation trick.

\paragraph{REINFORCE} The REINFORCE gradient estimator, also known as the \emph{score function estimator}, is proposed by using the log-derivative trick: $\nabla_{\pmb{\theta}}\log q_{\pmb{\theta}}(\mathbf{v}) = \frac{\nabla_{\pmb{\theta}} q_{\pmb{\theta}}(\mathbf{v})}{q_{\pmb{\theta}}(\mathbf{v})}$ which comes as a result of the chain rule \citep{glynn1990reinforce,williams1992reinforce}. Formally, the partial derivatives of parameters $\pmb{\theta}$ are given by the REINFORCE estimator
\begin{align*}
    \widehat{\nabla_{\pmb{\theta}}\ell}^{SCORE} = \ell \cdot (\nabla_{\pmb{\theta}}\log q_{\pmb{\theta}})
\end{align*}
It is an unbiased estimator of $\nabla_{\pmb{\theta}}\mathbb{E}_{q_{\pmb{\theta}}}[\ell]$, and only requires that one is able to evaluate $q_{\pmb{\theta}}$ and sample from it. The latter requirement is not restrictive and easily achieved in most settings since the form of $q_{\pmb{\theta}}$ is often assumed as part of model specification such as variational inference. Despite these desirable properties, it is well-known that the REINFORCE estimator suffers from high variance \citep{greensmith2001variance,xu2019variance}.

\paragraph{Reparameterisation Trick} The reparameterisation trick is compatible with random variables that have a location-scale parameterisation or have tractable inverse cumulative distribution functions (CDFs)\citep{price1958gaussian,glasserman2003monte,kingma2014vae,rezende2014stochastic,titsias2014doubly,figurnov2018implicit,jankowiak2018pathwise}. For Gaussian random variables, they are reparameterisable using a location-scale transformation which lets us write
\begin{align*}
    \mathbf{v} \sim \mathcal{N}(\pmb{\mu}, \pmb{\Sigma}) \overset{d}{=}g(\pmb{\epsilon}, \pmb{\mu}, \mathbf{V}) = \pmb{\mu} + \mathbf{V} \pmb{\epsilon}
\end{align*}
where $\mathbf{V} \in \R^{n \times n}$, $\pmb{\Sigma} = \mathbf{V}\mathbf{V}^{\top}$ and $\pmb{\epsilon} \sim q_0(\pmb{\epsilon}) =\mathcal{N}(\mathbf{0}, \mathbf{I})$. The latter expression for $\mathbf{v}$ is compatible with automatic differentiation and enables gradient-based updates to $\pmb{\mu}$ and $\mathbf{V}$. Typically, $\mathbf{V}$ is a diagonal matrix parameterised by a vector of variances $\pmb{\tau} = (\sigma_{1}^2, \dots, \sigma_{n}^2)$. That is, $\mathbf{V}=\pmb{\Sigma}^{\frac{1}{2}}$ with $\pmb{\Sigma} = \texttt{diag}(\pmb{\tau})$, so we have $\pmb{\theta} = \{\pmb{\mu}, \pmb{\tau}\}$ as our training parameters. Formally, we can write $\ell(g(\pmb{\epsilon}, \pmb{\theta}))$ and apply the chain rule to yield the reparameterisation gradient estimator
\begin{align}
\label{equation:reparam_gradient_estimator}
    \widehat{\nabla_{\pmb{\theta}}\ell}^{RT} = \left(J_{\ell}(\mathbf{v}) \cdot 
    \begin{bmatrix}
    \begin{array}{c|c}
         \mathbf{I}_n & \frac{1}{2}\pmb{\Sigma}^{-\frac{1}{2}} \odot (\mathbf{1}_n\pmb{\epsilon}^{\top})
    \end{array}
    \end{bmatrix}
    \right)^{\top}
\end{align}
where $J_{\ell}$ is the Jacobian of $\ell$ with respect to $\mathbf{v}$, and $[A|B]$ denotes a partitioned matrix with block matrices $A$ and $B$. The terms in the matrix are derived from the Jacobians $J_{g}(\pmb{\mu}) = \mathbf{I}_n$ and $J_{g}(\pmb{\tau}) = \frac{1}{2}\pmb{\Sigma}^{-\frac{1}{2}} \odot (\mathbf{1}_n\pmb{\epsilon}^{\top})$. Equation \ref{equation:reparam_gradient_estimator} is an unbiased estimator of $\nabla_{\pmb{\theta}}\mathbb{E}_{q_{\pmb{\theta}}}[\ell]$ due to the equivalence in distribution achieved by the reparameterisation trick. In the context of probabilistic modelling, the reparameterisation trick is often applied to individual scalar inputs of a decoder within a variational autoencoder (VAE) or individual weights in a Bayesian neural network (BNN). The latter may also be referred to as the \emph{global} reparameterisation trick.

\paragraph{Local Reparameterisation Trick} The local reparameterisation trick was proposed for NNs where the \emph{weights of linear layers} are independent Gaussian random variables \citep{kingma2015local}. Given an input $\mathbf{x} \in \R^n$, the pre-activations of these linear layers are given by
\begin{align*}
    \mathbf{z} =
    \begin{bmatrix}
    \mathbf{x}^{\top} g(\pmb{\epsilon}^{(1)}, \pmb{\theta}^{(1)}) \\
    \vdots \\
    \mathbf{x}^{\top} g(\pmb{\epsilon}^{(m)}, \pmb{\theta}^{(m)})\\
    \end{bmatrix}
    \in \R^{m}
\end{align*}
where $\pmb{\epsilon}^{(i)}$ and $\pmb{\epsilon}^{(j)}$ are independent for all $i\neq j$, and they admit a factorised Gaussian distribution $\Tilde{q}_{\mathbf{z}} = \prod_{i=1}^{m}\Tilde{q}_{z_i}$ where each pre-activation $z_i \sim \Tilde{q}_{z_i}$. Instead of sampling $g$ and computing $\mathbf{x}^{\top} g$, it is more efficient to \emph{locally} apply the reparameterisation trick to directly sample scalar pre-activations 
\begin{align}
\label{equation:local_reparam_trick}
    z_i \sim \mathcal{N}\left(\sum_{j=1}^{n}x_{j}\mu_{j}^{(i)}, \sum_{j=1}^{n}x_{j}^2\left(\sigma_{j}^{(i)}\right)^2\right) \overset{d}{=} \sum_{j=1}^{n}x_{j}\mu_{j}^{(i)} + \left(\sum_{j=1}^{n}x_{j}^2\left(\sigma_{j}^{(i)}\right)^2\right)^{\frac{1}{2}} \xi_i 
\end{align} 
where $\pmb{\mu}^{(i)}$ and $\pmb{\tau}^{(i)} = \left(\big(\sigma_{1}^{(i)}\big)^2, \dots, \big(\sigma_{n}^{(i)}\big)^2\right)$ are the \emph{global} mean and variance parameters of the Gaussian variables used to compute $z_i$ respectively, $\pmb{\theta}^{(i)} = \left\{\pmb{\mu}^{(i)}, \pmb{\tau}^{(i)}\right\}$, and $\xi_i \sim q_0^{(i)} = \mathcal{N}(0,1)$. The latter expression in Equation \ref{equation:local_reparam_trick} is known as the local reparameterisation trick as it changes the parameterisation of each $z_i$ as a scalar function of $\pmb{\epsilon}^{(i)}$ and $\pmb{\theta}^{(i)}$. Formally, we can write $\ell(g(\pmb{\epsilon}, \pmb{\theta})) = \Tilde{\ell}\left(z_1\left(\pmb{\epsilon}^{(1)}, \pmb{\theta}^{(1)}\right),\dots,z_m\left(\pmb{\epsilon}^{(m)}, \pmb{\theta}^{(m)}\right)\right)$ where $\Tilde{\ell}:\R^m \to \R$ denotes functions applied to $\mathbf{z}$ to compute the loss $\ell$ (i.e. non-linearities and upper NN layers), and apply the chain rule to yield the local reparameterisation gradient estimator for each $\pmb{\theta}^{(i)}$
\begin{align}
\label{equation:local_reparam_gradient_estimator}
    \widehat{\nabla_{\pmb{\theta}^{(i)}}\ell}^{LRT} = \left(\frac{\partial \Tilde{\ell}}{\partial z_i} \cdot 
    \begin{bmatrix}
    \begin{array}{c|c}
        \mathbf{x}^{\top} & \frac{1}{2}\left(\sum_{j=1}^{n}x_{j}^2\left(\sigma_{j}^{(i)}\right)^2\right)^{-\frac{1}{2}} \xi_i \left(\mathbf{x} \odot \mathbf{x}\right)^{\top}
    \end{array}
    \end{bmatrix}
    \right)^{\top}
\end{align}
where $\odot$ denotes element-wise product of matrices. The terms in the matrix are derived from the Jacobians $J_{z_i}(\pmb{\mu}^{(i)}) = \mathbf{x}^{\top}$ and $J_{z_i}(\pmb{\tau}^{(i)}) = \frac{1}{2}\left(\sum_{j=1}^{n}x_{j}^2\left(\sigma_{j}^{(i)}\right)^2\right)^{-\frac{1}{2}} \xi_i \left(\mathbf{x} \odot \mathbf{x}\right)^{\top}$. Equation \ref{equation:local_reparam_gradient_estimator} is an unbiased estimator of $\nabla_{\pmb{\theta}}\mathbb{E}_{q_{\pmb{\theta}^{(i)}}}[\ell]$ due to the equivalence in distribution achieved by the local reparameterisation trick. A notable application of the local reparameterisation trick is when performing mean-field variational inference within the linear layer of a BNN. It has been empirically observed that Equation \ref{equation:local_reparam_gradient_estimator} has lower variance than Equation \ref{equation:reparam_gradient_estimator} \citep{kingma2015local}. In Theorem \ref{theorem:local_reparam_as_r2g2}, we show that the local reparameterisation gradient estimator is equivalent to applying Rao-Blackwellisation to the reparameterisation gradient estimator and thereby enjoys variance reduction benefits of Rao-Blackwellisation.

\section{R2-G2 Gradient Estimator}
\label{section:r2g2}
    The reparameterisation trick can be seen as a procedure that couples the specification of a probabilistic model with an expression for gradients. In turn, assumptions on the former would restrict the applicability of the latter. This highlights the disadvantage of the local reparameterisation trick: the probabilistic model and gradients induced by the local reparameterisation trick is not suitable for general settings where the pre-activations $\mathbf{z}$ do not admit a factorised Gaussian distribution (i.e. the covariance matrix of $\mathbf{z}$ is not diagonal). On the other hand, the local reparameterisation trick has been empirically shown to enjoy lower variance of gradients than the global reparameterisation trick. 

To get the best of both worlds, we seek a reparameterisation gradient estimator that is general-purpose and enjoys reduced variance. To this end, we present our contribution: the \textbf{R}ao-Blackwellised \textbf{R}eparameterisation \textbf{G}radient Estimator for \textbf{G}aussian random variables, coined the R2-G2 estimator, as the Rao-Blackwellisation of the reparameterisation gradient estimator by conditioning on the realisation of multivariate Gaussian vectors resultant from linear transformations of Gaussian vectors. We first describe the Rao-Blackwellisation of the reparameterisation gradient estimator. We then provide the analytical form of the R2-G2 estimator and a summary of key properties of the R2-G2 estimator and its connection to existing methods. In particular, we show that the local reparameterisation gradient estimator is an instance of the R2-G2 estimator. We then conclude with a practical implementation to bypass costly matrix inversion operations by reformulating matrix-vector products as the solution to a quadratic optimisation problem. We defer all proofs to Appendix \ref{appendix:r2g2_key_properties}.

\begin{table}
  \caption{Examples of linear maps applied to Gaussian variables in probabilistic neural networks.}
  \label{table:application_examples}
  \centering
  \begin{tabular}{lll}
    \toprule
    Application & $\mathbf{W}$ & $\mathbf{V} \pmb{\epsilon}$ \\
    \midrule
    BNN & Outputs of a previous hidden layer & Gaussian weights in a linear layer \\
    BNN & Vectorised patches of an image & Gaussian weights in a convolutional layer \\
    VAE & Linear layer after reparameterisation & Gaussian latent variables  \\
    \bottomrule
  \end{tabular}
\end{table}

\subsection{Rao-Blackwellisation of the Reparameterisation Gradient Estimator}
\label{subsection:rao_blackwellisation_reparam}
The idea behind our Rao-Blackwellisation scheme is to condition on linear transformations of Gaussian variables since conditional Gaussian distributions can be described analytically. This is further motivated by the observation that we can decompose loss evaluations in NNs as
\begin{align}
\label{equation:loss_decomposition}
    \ell(g(\pmb{\epsilon}, \pmb{\theta})) = (\Tilde{\ell} \circ \mathbf{W}) (g(\pmb{\epsilon}, \pmb{\theta}))
\end{align}
where $\mathbf{W}:\R^n \to \R^m$ is a linear map and $\mathbf{W} \cdot g(\pmb{\epsilon}, \pmb{\theta})$ are pre-activations. This generalises the idea of the local reparameterisation trick where $\mathbf{W}$ is a row vector (i.e. $\mathbf{W}=\mathbf{x}^{\top}$) and inputs of $\Tilde{\ell}$ are $m$ scalar pre-activations. In deep learning, $\mathbf{W}$ frequently appears as hidden layers within NNs. A list of common linear transformations in probabilistic NNs is given in Table \ref{table:application_examples}. Applying the chain rule to Equation \ref{equation:loss_decomposition} gives an alternative expression of the reparameterisation gradient estimator
\begin{align}
\label{equation:reparam_gradient_estimator_decomposed}
    \widehat{\nabla_{\pmb{\theta}}\ell}^{RT} = \left(J_{\Tilde{\ell}}(\mathbf{W} \cdot g) \cdot \mathbf{W} \cdot  
    \begin{bmatrix}
    \begin{array}{c|c}
         \mathbf{I}_n & \frac{1}{2}\pmb{\Sigma}^{-\frac{1}{2}} \odot (\mathbf{1}_n\pmb{\epsilon}^{\top})
    \end{array}
    \end{bmatrix}
    \right)^{\top}
\end{align}
where $J_{\Tilde{\ell}}$ is the Jacobian of $\Tilde{\ell}$. With Equation \ref{equation:reparam_gradient_estimator_decomposed} in hand, we now present the R2-G2 estimator\footnote{Equations \ref{equation:r2g2_expectation} and \ref{equation:r2g2_closed} in Definition \ref{definition:r2g2} only hold when $\mathbf{v}$ are independent Gaussian variables. A more general expression can be derived when independence does not hold.}.
\begin{definition}[R2\text{-}G2]
\label{definition:r2g2}
The R2\text{-}G2 gradient estimator is given by
\begin{align}
\label{equation:r2g2_expectation}
    \widehat{\nabla_{\pmb{\theta}}\ell}^{R2\text{-}G2} = \mathbb{E}_{\Tilde{q}_0}\left[\widehat{\nabla_{\pmb{\theta}}\ell(\pmb{\epsilon})}^{RT}\right] &= \bigg(J_{\Tilde{\ell}}(\mathbf{W} \cdot g) \cdot \mathbf{W} \cdot \mathbb{E}_{\Tilde{q}_0}
    \begin{bmatrix}
    \begin{array}{c|c}
         \mathbf{I}_n & \frac{1}{2}\pmb{\Sigma}^{-\frac{1}{2}} \odot (\mathbf{1}_n\pmb{\epsilon}^{\top})
    \end{array}
    \end{bmatrix}
    \bigg)^{\top}
\end{align}
where $\pmb{\epsilon}| \mathbf{W} \cdot g = \mathbf{z} \sim \Tilde{q}_0$ with mean $\pmb{\epsilon}^*=\mathbf{A}^{\top}\left(\mathbf{A}\mathbf{A}^{\top}\right)^{\dagger}(\mathbf{z}-\mathbf{W}\pmb{\mu}) = \mathbf{A}^{\top}\left(\mathbf{A}\mathbf{A}^{\top}\right)^{\dagger}\mathbf{A}\pmb{\epsilon}$ with $\mathbf{A} = \mathbf{W}\mathbf{V}$. By linearity of expectations, it also admits a closed-form expression
\begin{align}
\label{equation:r2g2_closed}
    \widehat{\nabla_{\pmb{\theta}}\ell}^{R2\text{-}G2} = \bigg(J_{\Tilde{\ell}}(\mathbf{W} \cdot g) \cdot \mathbf{W} \cdot 
    \begin{bmatrix}
    \begin{array}{c|c}
         \mathbf{I}_n & \frac{1}{2}\pmb{\Sigma}^{-\frac{1}{2}} \odot (\mathbf{1}_n(\pmb{\epsilon}^*)^{\top})
    \end{array}
    \end{bmatrix}
    \bigg)^{\top}.
\end{align}
\end{definition}
The R2-G2 estimator is a single-sample gradient estimator as it is a conditional expectation of the reparameterisation gradient estimator formed by conditioning on a single sample of the vector of pre-activations $\mathbf{z}$, making it the Rao-Blackwellisation \citep{blackwell1947conditional, rao1992information} of the reparameterisation gradient estimator from Equation \ref{equation:reparam_gradient_estimator_decomposed}. The law of total variance lets us deduce that the R2-G2 estimator has lower variance since it swaps the random matrix in Equation \ref{equation:reparam_gradient_estimator_decomposed} with its conditional expectation. We conclude our description of the R2-G2 estimator with its key properties, namely unbiasedness and enjoying lower variance than the reparameterisation gradient estimator.
\begin{proposition}
\label{proposition:r2g2_key_properties}
Denote $\mathbf{z} \sim q_{\mathbf{z}} = \mathcal{N}\left(\mathbf{W} \cdot \pmb{\mu}, \mathbf{A} \mathbf{A}^{\top} \right)$. Then we have
\begin{align*}
\mathbb{E}_{q_{\mathbf{z}}}\left[\widehat{\nabla_{\pmb{\theta}}\ell}^{R2\text{-}G2}\right] = \nabla_{\pmb{\theta}}\mathbb{E}_{q_{\pmb{\theta}}}[\ell]
\end{align*}
and
\begin{align*}
\mathbb{E}_{q_{\mathbf{z}}}\left[\left\|\widehat{\nabla_{\pmb{\theta}}\ell}^{R2\text{-}G2} -  \nabla_{\pmb{\theta}}\mathbb{E}_{q_{\pmb{\theta}}}[\ell]\right\|^2\right] \leq \mathbb{E}_{q_0}\left[\left\|\widehat{\nabla_{\pmb{\theta}}\ell}^{RT} -  \nabla_{\pmb{\theta}}\mathbb{E}_{q_{\pmb{\theta}}}[\ell]\right\|^2 \right].
\end{align*}
\end{proposition}

\subsection{Computation of Rao-Blackwellisation scheme as a least-squares problem}
The R2-G2 estimator exploits the analytical form of the conditional Gaussian distribution. This is done by either sampling from the conditional Gaussian distribution $\Tilde{q}_0$ within Equation \ref{equation:r2g2_expectation} or directly using the mean of the conditional Gaussian distribution $\pmb{\epsilon}^{*}$ in place of $\pmb{\epsilon}$ within Equation \ref{equation:r2g2_closed}. Equation \ref{equation:r2g2_expectation} requires computing the Cholesky factor of the covariance matrix of $\Tilde{q}_0$, which incurs a $\mathcal{O}(m^3)$ computational cost and a $\mathcal{O}(m^2)$ storage cost. The latter makes it impractical for training NNs as each gradient descent step would require instantiating the matrix $\mathbf{A}\mathbf{A}^{\top}$. This makes Equation  \ref{equation:r2g2_closed} more desirable as long as $\pmb{\epsilon}^{*}$ is computed and stored efficiently. To do so, we consider the linear system
\begin{equation}
\label{equation:normal_equation}
\begin{aligned}
    \mathbf{A}\mathbf{A}^{\top}\pmb{\beta} = \mathbf{A}\pmb{\epsilon}.
\end{aligned}
\end{equation}
It can be verified that any solution $\pmb{\beta}^{*}$ to Equation \ref{equation:normal_equation} satisfies $\pmb{\epsilon}^{*} = \mathbf{A}^{\top} \pmb{\beta}^{*}$ (see Appendix \ref{appendix:verifying_conditional_mean}).

\paragraph{Least-squares characterisation of Rao-Blackwellisation}
Observe that Equation \ref{equation:normal_equation} is a normal equation that describes the first-order optimality condition of the quadratic optimisation problem
\begin{align*}
    \underset{\pmb{\beta} \in \mathbb{R}^m}{\text{minimise}} \quad \frac{1}{2}\|\pmb{\epsilon} - \mathbf{A}^{\top}\pmb{\beta}\|_2^2 \Leftrightarrow \underset{\pmb{\beta} \in \mathbb{R}^m}{\text{maximise}} \quad \exp\left(-\frac{1}{2}\|\pmb{\epsilon} - \mathbf{A}^{\top}\pmb{\beta}\|_2^2\right).
\end{align*}
We can interpret computing a solution $\pmb{\beta}^{*}$ as fitting a single-sample multivariate linear model $\pmb{\epsilon} | \mathbf{A}$
\begin{align*}
    \pmb{\epsilon} &= \mathbf{A}^{\top}\pmb{\beta} + \pmb{\delta}
\end{align*}
where $\pmb{\beta} \in \R^m$ and $\pmb{\delta} \sim \mathcal{N}(\mathbf{0}_n, \mathbf{I}_n)$. Under this linear model, the goal is to solve for $\pmb{\beta}^{*}$ by using covariance parameters $\mathbf{A}$ as \emph{predictors} of the noise $\pmb{\epsilon}$. With $\pmb{\beta}^{*}$ computed, we can then compute 
\begin{equation}
\label{equation:least_squares_epsilon}
\begin{aligned}
    \pmb{\epsilon}^{*} = \mathbf{A}^{\top} \pmb{\beta}^{*} = \mathbf{A}^{\top}\left(\mathbf{A}\mathbf{A}^{\top}\right)^{\dagger}\mathbf{A} \pmb{\epsilon}
\end{aligned}
\end{equation}
which corresponds to the expression of the mean of a conditional Gaussian distribution  (see Lemma \ref{lemma:mvn_properties}). In other words, computing the R2-G2 estimator is inherently solving a least squares problem and applying a linear transformation. Intuitively, the former is the mechanism giving variance reduction. In the language of linear models, the observed noise $\pmb{\epsilon}$ is projected to the \emph{fitted} noise $\pmb{\epsilon}^{*}$ which has minimal Euclidean norm.

\paragraph{Iterative solver}
\label{subsection:r2g2_as_least_squares}
\begin{wrapfigure}[8]{r}{0.52\textwidth}
\vspace{-2.5em}
\begin{minipage}{0.52\textwidth}
\begin{algorithm}[H]
   \caption{Forward Pass with R2-G2 Gradients.}
   \label{algorithm:linear_with_r2g2_gradients}
\begin{algorithmic}
   \STATE {\bfseries Input:} matrix $\mathbf{A}$, noise vector $\pmb{\epsilon}$.
   \STATE Compute $\mathbf{z} = \mathbf{A}\pmb{\epsilon}$.
   \STATE Compute $\pmb{\beta}^{*} = \texttt{conjugate\_gradient}(\mathbf{A}, \mathbf{z})$.
   \STATE Compute $\pmb{\epsilon}^{*} = \mathbf{A}^{\top}\pmb{\beta}^{*}$.
   \STATE Compute $\mathbf{z}^{*} = \mathbf{A}\pmb{\epsilon}^{*}$.
   \STATE {\bfseries Output:} $\texttt{stop\_gradient}\big(\mathbf{z} - \mathbf{z}^{*}) + \mathbf{z}^{*}$.
\end{algorithmic}
\end{algorithm}
\end{minipage}
\end{wrapfigure}
By defining functions that compute matrix-vector products with the matrix $\mathbf{A}\mathbf{A}^{\top}$, we can use the conjugate gradient algorithm to solve Equation \ref{equation:normal_equation} for $\pmb{\beta}^{*}$. The conjugate gradient algorithm is an iterative method that terminates in at most $\texttt{rank}(\mathbf{A})\leq m$ iterations \citep{kaasschieter1988preconditioned, nocedal1999numerical,hayami2018convergence}, and only requires storing the updated solution at each iteration. While the former implies that the worst-case computational costs remains at $\mathcal{O}(m^3)$, the latter implies the storage cost is reduced to $\mathcal{O}(m)$ and makes it practical to compute $\pmb{\beta}^{*}$. In our setting, the worst-case computational costs of  $\mathcal{O}(m^3)$ can be reduced since the structure of $\mathbf{A}$ is known (see Appendix \ref{appendix:comparison_computational_cost}). With $\pmb{\beta}^{*}$ in hand, we can compute $\pmb{\epsilon}^{*}$ and modify forward computations to use the R2-G2 gradient estimator for backpropagation\footnote{In Algorithm \ref{algorithm:linear_with_r2g2_gradients}, computing $\mathbf{z}^{*}$ ensures automatic differentiation calculates Equation \ref{equation:r2g2_closed} for backpropagation, while the output $\texttt{stop\_gradient}\big(\mathbf{z} - \mathbf{z}^{*}) + \mathbf{z}^{*}$ ensures the forward computation is still performed with $\mathbf{z}$.}, by implementing Algorithm \ref{algorithm:linear_with_r2g2_gradients} in deep learning frameworks such as PyTorch \citep{pytorch2019}.

\subsection{Connections to related work}
\paragraph{Local Reparameterisation Gradient Estimator} By setting $\mathbf{V}^{(i)}=\left(\texttt{diag}(\pmb{\tau}^{(i)})\right)^{\frac{1}{2}}$, $\mathbf{W} = \mathbf{x}^{\top}$ and the conditional distribution $\pmb{\epsilon}^{(i)} | z_i \sim \Tilde{q}^{(i)}_0$ for $i=1,\dots,m$ within Definition \ref{definition:r2g2}, we are able to show that the local reparameterisation gradient estimator is equivalent to the R2-G2 estimator. Theorem \ref{theorem:local_reparam_as_r2g2} presents this equivalence and formalises the empirical variance reduction of the local reparameterisation trick observed in the experiments of \cite{kingma2015local}, as an instance of Rao-Blackwellisation and the R2-G2 estimator. Formally, it shows the local reparameterisation trick is equivalent to using pre-activation samples for forward computations and a Rao-Blackwellised \emph{global} reparameterisation gradient estimator to update parameters in linear layers.
\begin{theorem}
\label{theorem:local_reparam_as_r2g2}
Suppose we have a BNN linear layer where weights are independent Gaussian random variables. That is, $\pmb{\theta}^{(i)} = \left\{\pmb{\mu}^{(i)}, \pmb{\tau}^{(i)}\right\}$ where $\pmb{\tau}^{(i)}=\left(\left(\sigma_{1}^{(i)}\right)^2, \dots, \left(\sigma_{n}^{(i)}\right)^2\right)$ for $i=1,\dots,m$. Then for each $i=1,\dots,m$, we have $\widehat{\nabla_{\pmb{\theta}^{(i)}}\ell}^{R2\text{-}G2} =\mathbb{E}_{\Tilde{q}^{(i)}_0}\left[\widehat{\nabla_{\pmb{\theta}^{(i)}}\ell}^{RT}\right] \overset{d}{=} \widehat{\nabla_{\pmb{\theta}^{(i)}}\ell}^{LRT}$ and 
\begin{align*}
\mathbb{E}_{\Tilde{q}_{z_i}}\left[\left\|\widehat{\nabla_{\pmb{\theta}^{(i)}}\ell}^{LRT} -  \nabla_{\pmb{\theta}^{(i)}}\mathbb{E}_{q_{\pmb{\theta}^{(i)}}}[\ell]\right\|^2\right] \leq \mathbb{E}_{q^{(i)}_0}\left[\left\|\widehat{\nabla_{\pmb{\theta}^{(i)}}\ell}^{RT} -  \nabla_{\pmb{\theta}^{(i)}}\mathbb{E}_{q_{\pmb{\theta}^{(i)}}}[\ell]\right\|^2 \right].
\end{align*}
\end{theorem}
As an analogy, we can view the way that R2-G2 generalises the local reparameterisation gradient estimator, in the way that square matrix inversion  generalises scalar inversion. Recall that Equation \ref{equation:local_reparam_trick} exploits the fact that the Gaussian pre-activations $\mathbf{z}$ of mean-field BNN linear layers have a diagonal covariance matrix \emph{by design}. For forward computations, this means each $z_i$ can be efficiently sampled with the reparameterisation trick by using the square root of the variance $\sigma_{z_i}^2 = \sum_{j=1}^{n}x_{j}^2\left(\sigma_{j}^{(i)}\right)^2$. The subtle benefit of reduced variance in gradients is due to the square root function having a derivative which matches its reciprocal (up to a scaling factor). This means that the variance $\sigma_{z_i}^2$ (i.e. a scalar) is \emph{inverted} when using the local reparameterisation gradient estimator. The R2-G2 estimator naturally extends the inversion of scalar variances to square covariance matrices, by calling the conjugate gradient algorithm, thereby making it suitable for other probabilistic models.

\paragraph{Variance reduction as Stochastic Linear Regression} Variance reduction through stochastic linear regression has previously been explored in the context of variational inference \cite{salimans2013fixed, salimans2014using, salimans2014implementing}. Specifically, variational inference is presented as fitting linear regression with the unnormalised log posterior as the dependent variable and the sufficient statistics of latent random variables as explanatory variables. The connection to variance reduction of gradients is then made by using \emph{multiple} samples to construct control variates that correlate with the gradient of the KL-divergence. This differs from our work as we present variance reduction of gradients as Rao-Blackwellisation through the R2-G2 estimator as a \emph{single-sample} gradient estimator, and explicate the connection to linear regression with the reparameterisation noise and covariance matrix parameters as the dependent and explanatory variables respectively.

\section{Experiments}
\label{section:experiments}
    \subsection{Protocol}
The R2-G2 estimator can be readily applied to any existing application of the reparameterisation trick for Gaussian variables. While this may appear restrictive, we note that Gaussian distributions often serve as a default prior or noise distribution in probabilistic ML tasks. To motivate our experiments, we note that the R2-G2 and reparameterisation gradient estimators would yield the same model when a \emph{large} number of gradient descent steps are taken, as the mean gradient of both estimators are the same. The focus of our experiments is during initial training, where only a \emph{small} number of gradient descent steps are taken. Examples of scenarios where this setting can be beneficial include pre-training to discover a better initialisation for a model, fine-tuning a pre-trained model, or when the amount of training is limited by a computation budget. 

We evaluate the benefits of initial training with the R2-G2 estimator for probabilistic models. We consider two standard tasks with variational Bayesian models that utilise a mean-field approximation of the posterior: image classification with BNNs and generative modelling with hierarchical VAEs. In our experiments, we provide numerical comparisons against the reparameterisation (RT) and local reparameterisation (LRT) gradient estimators, where permissible. We do not compare against the REINFORCE estimator as it is well-known that its high variance makes optimisation difficult. Unless stated otherwise, we compute pre-activations in the same way as the reparameterisation trick (i.e. sampling $g(\pmb{\epsilon}, \pmb{\theta})$ and computing $\mathbf{W}\cdot g$). We defer the full details of our experiments to Appendix \ref{appendix:experiment_details}. 

\subsection{Image classification with Bayesian Neural Networks}
\label{experiments:image_classification_bnns}
\begin{table}[t]
\caption{Log-likelihoods and classification accuracies (\%) of BNNs using the R2-G2, Reparameterisation (RT) and Local Reparameterisation (LRT) estimators over 5 runs. Higher is better. Error bars denote $\pm 1.96$ standard errors $(\sigma / \sqrt{5})$ over 5 runs. See text for details.}
\label{table:bnn_results}
\begin{center}
\begin{tabular}{lccccc}
\toprule
& & \multicolumn{2}{c}{Log-likelihood} & \multicolumn{2}{c}{Accuracy} \\
\cmidrule(lr){3-4} \cmidrule(lr){5-6}
Dataset & Estimator & Train & Test & Train & Test \\ 
\midrule
MNIST & R2-G2 & $\mathbf{-3.00
 \pm 0.00}$ & $\mathbf{-3.06
 \pm 0.00}$ & $99.81
 \pm 0.03$ & $\mathbf{98.00
 \pm 0.08}$ \\
& LRT & $\mathbf{-3.00
 \pm 0.00}$ & $\mathbf{-3.06
 \pm 0.00}$ & $99.81
 \pm 0.04$ & $97.99
 \pm 0.07$\\
& RT & $\mathbf{-3.00
 \pm 0.00}$ & $-3.07
 \pm 0.00$ & $\mathbf{99.85
 \pm 0.04}$ & $\mathbf{98.00
 \pm 0.11}$\\
\midrule
CIFAR-10 & R2-G2 & $\mathbf{-3.83  \pm 0.01}$ & $\mathbf{-3.87
 \pm 0.01}$ & $\mathbf{70.08 \pm 0.43}$ & $67.97 \pm 0.66$\\
& RT & $-3.84  \pm 0.01$ & $-3.88 \pm 0.02$ & $69.76 \pm 0.46$ & $\mathbf{67.98 \pm 0.88}$\\
\bottomrule
\end{tabular}
\end{center}
\end{table}

We consider fully stochastic BNNs for image classification tasks on two standard benchmark datasets: MNIST \citep{lecun2010mnist} and CIFAR-10 \citep{krizhevsky2009learning}. In this setting, the input data $\mathcal{D}$ is a dataset and the latent variables $\mathbf{v}$ are weights (i.e. $\mathbf{v}$ are global variables). To enable training with minibatches, we use the stochastic approximation of the variational lower bound
\begin{align*}
    \log p(y | \mathbf{x}) > \mathbb{E}_{q_{\pmb{\theta}}(\mathbf{v}| \mathcal{D})} \left[\log \left(\frac{p(\mathbf{v})}{q_{\pmb{\theta}}(\mathbf{v}| \mathcal{D})}\right) + \frac{N}{B}\sum_{i=1}^{B}\log p(y^{(i)} | \mathbf{x}^{(i)}, \mathbf{v})  \right]
\end{align*}
where $N$ is the size of the dataset $\mathcal{D}$ and $\left\{(\mathbf{x}^{(i)}, y^{(i)})\right\}_{i=1}^{B} \subset \mathcal{D}$ is a minibatch of size $B < N$. We use the standard train and test splits of both datasets. See Table \ref{table:bnn_results} for classification accuracies and log-likelihoods reported on the train and test sets of each dataset.

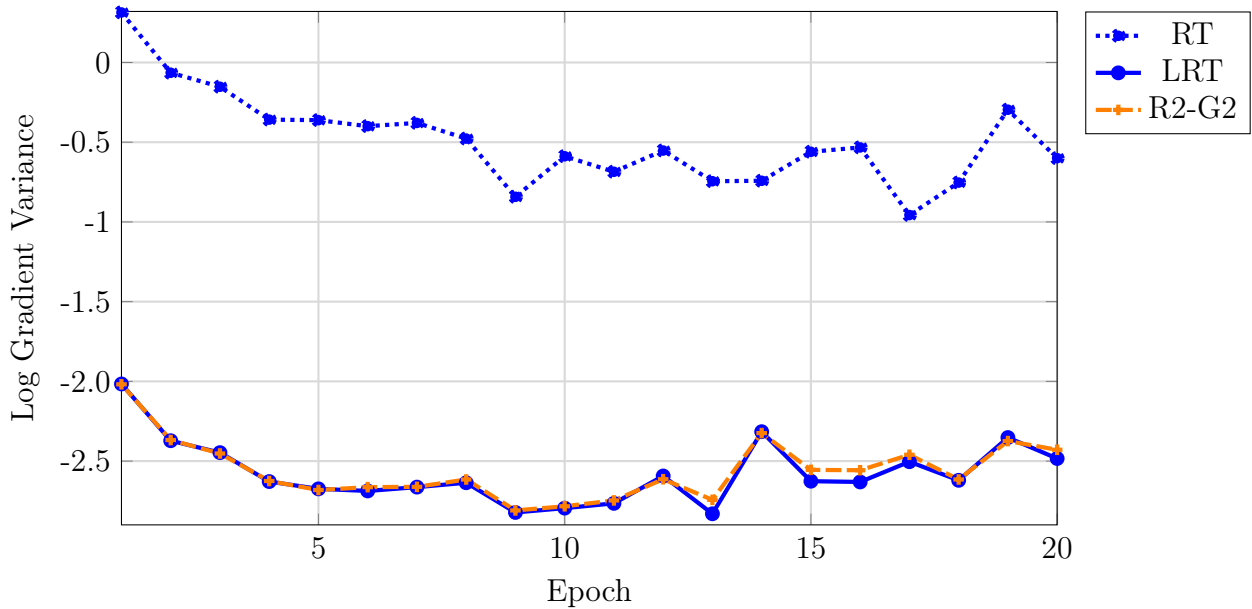
\begin{figure*}[h]
  \centering
  \begin{tikzpicture}
  \begin{axis}[
      width=0.8*\textwidth,
      height=0.6*0.8*\textwidth,
      grid=major, 
      grid style={thick,gray!30},
      xlabel= Epoch, 
      ylabel= Log Gradient Variance,
      legend pos= outer north east,
      x tick label style={ultra thick, anchor=north},
      xmin=1, xmax=20,
      ymin=-2.9, ymax=0.32,
      xtick={5, 10, 15, 20},
      ytick={-2.5, -2.0, -1.5, -1, -0.5, 0},
      yticklabels={-2.5, -2.0, -1.5, -1, -0.5, 0},
      scaled x ticks = false, 
      x tick label style={/pgf/number format/.cd, fixed, fixed zerofill, int detect, 1000 sep={}}
    ]
    \addplot [ultra thick, dotted, blue, mark=*,text mark as node=true] 
    table[x=epoch,y=log_gradient_variance,col sep=comma, discard if not={estimator}{RT}, discard if not={layer}{top}]{results/bnn_mnist_grad_var.csv};
    \addplot [ultra thick, blue, mark=*,text mark as node=true] table[x=epoch,y=log_gradient_variance,col sep=comma, discard if not={estimator}{LRT}, discard if not={layer}{top}]{results/bnn_mnist_grad_var.csv};
    \addplot [ultra thick, dash pattern=on 6pt off 2pt, orange, mark=+,text mark as node=true] table[x=epoch,y=log_gradient_variance,col sep=comma, discard if not={estimator}{R2-G2}, discard if not={layer}{top}]{results/bnn_mnist_grad_var.csv};

    \legend{
        RT,
        LRT,
        R2-G2,
    }
  \end{axis}
\end{tikzpicture}
  \caption{Log average gradient variance v.s. epoch for the top layer of a Bayesian MLP trained on \emph{MNIST} over 5 runs. We compare the variance of gradients when training using the reparameterisation (RT), local reparameterisation (LRT) and R2-G2 estimators.}
  \label{figure:bnn_mnist_grad_var_top}
\end{figure*}

For MNIST, we used the two-layer multi-layer perceptron (MLP) architecture from \cite{srivastava2014dropout}, which consists of two hidden layers of 1024 \texttt{ReLU} units. Each estimator is applied to all linear layers. For experiments with the R2-G2 estimator, we compute pre-activations in the same way as the local reparameterisation trick (i.e. sampling pre-activations directly). Our results illustrate that accuracies and log-likelihoods for the R2-G2 and LRT estimators match exactly. In Figures \ref{figure:bnn_mnist_grad_var_top} and \ref{figure:bnn_mnist_grad_var_bottom}, the variance of gradients for the R2-G2 and LRT estimators also coincide and are much lower than those of the RT estimator. These observations empirically supports Theorem \ref{theorem:local_reparam_as_r2g2}: the LRT estimator is an instance of the R2-G2 estimator and Rao-Blackwellisation. In other words, the local reparameterisation trick is equivalent to directly sampling pre-activations in forward computations and using a Rao-Blackwellised RT estimator in backpropagation.

For CIFAR-10, we used the VGG-11 architecture described in \cite{simonyan2015vgg}, where each estimator is applied to the last four convolutional layers. For the last four convolutional layers, we do not include the local reparameterisation trick as a benchmark since sampling \emph{scalar} pre-activations directly would ignore dependencies induced by sharing weights in a convolutional layer (i.e. \emph{additionally assumes} $\mathbf{A}\mathbf{A}^{\top}$ is diagonal). The RT estimator is applied to the first four convolutional layers. To limit the effect of gradient variances from linear layers on gradient variances of convolutional layers, we applied the LRT estimator to all linear layers. 

Across both datasets, we found that initial training with the R2-G2 estimator yields higher log-likelihoods than the RT estimator while enjoying similar levels of accuracy. These results extend the benefits of training with Rao-Blackwellised gradients from Bayesian MLPs to Bayesian CNNs.

\subsection{Generative modelling with Hierarchical Variational Autoencoders}
\label{experiments:generative_modelling_vaes}
\begin{table}[t]
\caption{Test variational lower bounds for hierarchical VAEs using the R2-G2 and Reparameterisation (RT) estimators. Higher is better. Error bars denote $\pm 1.96$ standard errors $(\sigma / \sqrt{5})$ over 5 runs. See text for details.}
\label{table:hvae_results}
\begin{center}
\begin{tabular}{lcccc}
\toprule
\# VAE Layers & Estimator & MNIST & Omniglot & Fashion-MNIST \\ 
\midrule
2 & R2-G2 & $\mathbf{-106.85 \pm 5.00}$ & $\mathbf{-129.80
 \pm 0.74}$ & $\mathbf{-240.23 \pm 0.64}$\\
& RT & $-107.64 \pm 8.46$ & $-131.48
 \pm 1.66$ & $-240.59 \pm 0.93$\\
\midrule
3 & R2-G2 & $\mathbf{-102.45 \pm 3.73}$ & $\mathbf{-134.95 \pm 2.26}$ & $\mathbf{-240.15 \pm 0.86}$\\
& RT & $-111.50 \pm 5.40$ & $-136.12 \pm 2.70$ & $-240.89 \pm 0.97$\\
\bottomrule
\end{tabular}
\end{center}
\end{table}
We consider VAEs for generative modelling tasks on three standard benchmark datasets: MNIST \citep{lecun2010mnist}, Omniglot \citep{lake2015omniglot} and Fashion-MNIST \citep{xiao2017fmnist}. We do not focus on comparisons of gradient variances due to architecture constraints of VAEs. Validating Proposition \ref{proposition:r2g2_key_properties} requires independent and identically distributed (i.i.d.) samples of $\mathbf{z} \sim \Tilde{q}_{\mathbf{z}}$ which requires computing a Cholesky factor in each iteration. This is computationally expensive and not representative of computations in VAEs. In practice, we compute $\mathbf{z} = \mathbf{W} \cdot g(\pmb{\epsilon}, \pmb{\theta})$ where $\mathbf{W}$ is the linear layer following reparameterisation, which does not enable sampling the full support of $\Tilde{q}_{\mathbf{z}}$. 

We found that training one-layer VAEs with the R2-G2 estimator did not guarantee performance gains (see Appendix \ref{appendix:1vae_experiments}), and surmise that reducing gradient variance hinders the learning of stable representations in this setting as only training of the encoder is affected by the R2-G2 estimator. An investigation of this phenomena is beyond the scope of this paper and we leave this for future work. 

We present the results of our experiments for two-layer and three-layer VAEs (i.e. hierarchical VAEs). In this setting, the input data $\mathcal{D}$ is an observation and the latent variables $\mathbf{v}$ is its latent representation (i.e. $\mathbf{v}$ are local variables). The objective is to maximise the variational lower bound on the log-likelihood
\begin{align*}
    \log p(\mathbf{x}) > \mathbb{E}_{q_{\pmb{\theta}}(\mathbf{v}^{(1)}| \mathcal{D}),\dots, q_{\pmb{\theta}}(\mathbf{v}^{(K)}| \mathcal{D})} \left[\log \left(\frac{1}{K} \sum_{i=1}^{K} \frac{p(\mathcal{D}, \mathbf{v}^{(i)})}{q_{\pmb{\theta}}(\mathbf{v}^{(i)}| \mathcal{D})}\right) \right]
\end{align*}
where $\mathcal{D}$ denotes input data, and $\{\mathbf{v}^{(i)}\}_{i=1}^{K}$ are vectors of latent Gaussian random variables. For training, we use a single sample ($K=1$), which is equivalent to variational inference. For testing, we use $5000$ samples ($K=5000$), which is equivalent to importance weighted variational inference, providing a tighter bound on the log-likelihood \citep{burda2016iwae}. For each $L$-layer VAE, we used a bottom-up variational posterior and top-down generative process with latent variables $\{\mathbf{v}_{i}\}_{i=1}^{L}$
\begin{align*}
    q(\{\mathbf{v}_{i}\}_{i=1}^{L}| \mathbf{x}) = q(\mathbf{v}_1| \mathbf{x}) \prod_{l=2}^{L} q(\mathbf{v}_{l}| \mathbf{v}_{l-1}, \mathbf{x}),\quad p(\mathbf{x}, \{\mathbf{v}_{i}\}_{i=1}^{L}) = p(\mathbf{x}| \{\mathbf{v}_{i}\}_{i=1}^{L}) \prod_{l=1}^{L-1} p(\mathbf{v}_{l}| \mathbf{v}_{l+1})
\end{align*}
with latent spaces of 50 units and each conditional distribution is parameterised by a MLP with two hidden layers of 200 $\texttt{tanh}$ units (see \cite{burda2016iwae,bauer2021gdreg}). We used a factorised Bernoulli likelihood, and factorised Gaussian variational posterior and prior. 

We applied the R2-G2 estimator as a single layer within the \emph{decoder} that applies reparameterisation and the linear transformation that follows it (i.e. $\mathbf{W}$ and $\mathbf{V}$ are decoder parameters). We used dynamic binarisation of all three datasets \citep{salakhutdinov2008dynamic}.  We used the standard train and test splits for MNIST and Fashion-MNIST, and the train and test split from \citep{burda2016iwae} for Omniglot.  See Table \ref{table:hvae_results} for test lower bounds on the log-likelihood of all datasets after 100,000 steps.

\begin{figure*}
  \begin{center}
    \begin{tikzpicture}
      \begin{axis}[
          width=0.45*\textwidth,
          height=0.8*0.4\textwidth,
          grid=major, 
          grid style={thick,gray!30},
          xlabel= Steps, 
          ylabel= ELBO,
          x tick label style={ultra thick, anchor=north},
          xmin=10000, xmax=100000,
          ymin=-160, ymax=-100,
          xtick={20000, 40000, 60000, 80000, 100000},
          xticklabels={20K, 40K, 60K, 80K, 100K},
          ytick={-150, -130, -110},
          yticklabels={-150, -130,  -110},
          scaled x ticks = false, 
          x tick label style={/pgf/number format/.cd, fixed, fixed zerofill, int detect, 1000 sep={}},
          legend columns=-1,
          legend entries={RT;, R2-G2},
          legend to name={mylegend_3vae_mnist},
          legend style={at={(0.5,-0.2)}, anchor=north,legend columns=2}
        ]
        \addplot [ultra thick, dotted, blue, mark=*,text mark as node=true] 
        table[x=steps,y=elbo,col sep=comma, discard if not={estimator}{RT}]{results/3vae_mnist_elbo_train.csv};
        \addplot [ultra thick, dash pattern=on 6pt off 2pt, orange, mark=+,text mark as node=true] table[x=steps,y=elbo,col sep=comma, discard if not={estimator}{R2-G2}]{results/3vae_mnist_elbo_train.csv};
      \end{axis}
    \end{tikzpicture}
    \begin{tikzpicture}
      \begin{axis}[
          width=0.45*\textwidth,
          height=0.8*0.4\textwidth,
          grid=major, 
          grid style={thick,gray!30},
          xlabel= Steps, 
          ylabel= ELBO,
          x tick label style={ultra thick, anchor=north},
          xmin=10000, xmax=100000,
          ymin=-160, ymax=-100,
          xtick={20000, 40000, 60000, 80000, 100000},
          xticklabels={20K, 40K, 60K, 80K, 100K},
          ytick={-150, -130, -110},
          yticklabels={-150, -130,  -110},
          scaled x ticks = false, 
          x tick label style={/pgf/number format/.cd, fixed, fixed zerofill, int detect, 1000 sep={}}
        ]
        \addplot [ultra thick, dotted, blue, mark=*,text mark as node=true] 
        table[x=steps,y=elbo,col sep=comma, discard if not={estimator}{RT}]{results/3vae_mnist_elbo_test.csv};
        \addplot [ultra thick, dash pattern=on 6pt off 2pt, orange, mark=+,text mark as node=true] table[x=steps,y=elbo,col sep=comma, discard if not={estimator}{R2-G2}]{results/3vae_mnist_elbo_test.csv};
      \end{axis}
    \end{tikzpicture}
    \\
    \ref*{mylegend_3vae_mnist}
  \end{center}
  
  \caption{Bounds on log-likelihood v.s. optimisation steps for a three-layer VAE trained on \emph{MNIST} over 5 runs. We compare the bounds on log-likelihoods when training using the reparameterisation (RT) and R2-G2 estimators. Training with the R2-G2 estimator improves bounds on log-likelihood on both the training set (left) and test set (right).}
  \label{figure:3vae_mnist_elbo}
\end{figure*}
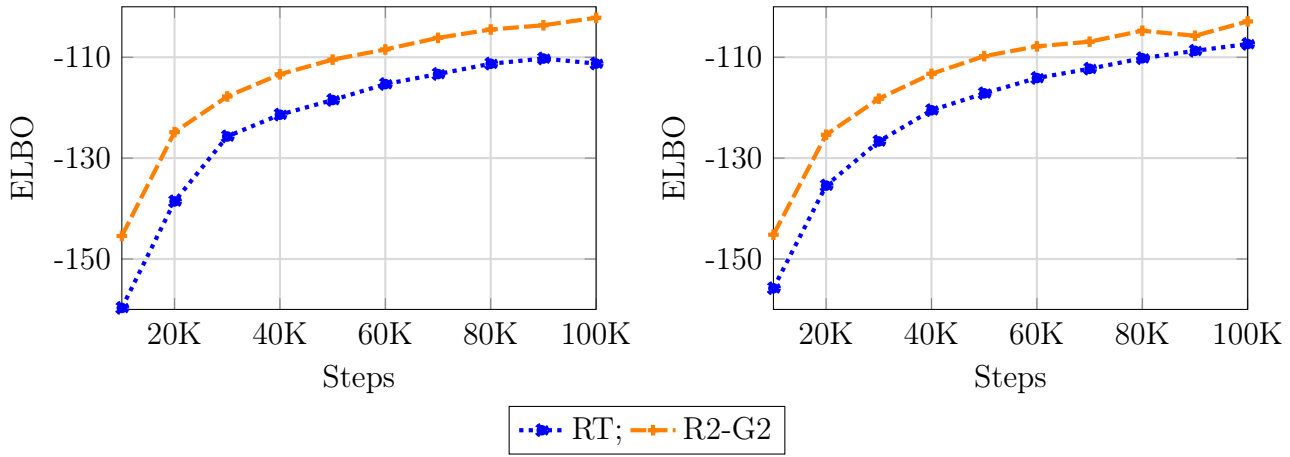

Across all datasets and hierarchical VAEs, we found that the R2-G2 estimator consistently yielded higher test ELBOs than the RT estimator. For three-layer VAEs, we saw substantial gains of 9.05 and 1.17 nats on MNIST and Omniglot respectively (see Figures \ref{figure:3vae_mnist_elbo} and \ref{figure:3vae_omniglot_elbo}). These results extend benefits of initial training with Rao-Blackwellised gradients, from Bayesian NNs to Hierarchical VAEs.

\section{Conclusion}
\label{section:conclusion}
    We have presented the R2-G2 estimator as a novel, general-purpose and single-sample gradient estimator for latent Gaussian random variables as the Rao-Blackwellisation of the reparameterisation gradient estimator. Our method is motivated by the widespread usage of Gaussian distributions in probabilistic ML and applications of Rao-Blackwellisation in gradient estimators for latent discrete variables \citep{liu2019rbreinforce,dong2020disarm,kool2020unordered,paulus2021gumbelrao}.

We theoretically and empirically show that the local reparameterisation trick is an instance of Rao-Blackwellisation and the R2-G2 estimator for linear layers of BNNs with independent weights. It is equivalent to sampling pre-activations in forward computations and updating parameters with a Rao-Blackwellised reparameterisation gradient estimator in backpropagation. This explicates the empirical evidence that the local reparameterisation trick reduces variance of gradients obtained by the global reparameterisation trick as the benefit of Rao-Blackwellisation stated by Theorem \ref{theorem:local_reparam_as_r2g2}. By casting the local reparameterisation gradient estimator as Rao-Blackwellised gradients for Bayesian MLPs, we showed that initial training with Rao-Blackwellised gradients also yield gains in performance for other models such as Bayesian CNNs and hierarchical VAEs. While performance gains were sometimes modest, they were consistent across models with multiple applications of the reparameterisation trick and particularly prominent in hierarchical VAEs. The main limitation of the R2-G2 estimator is the computational cost from solving Equation  \ref{equation:normal_equation} and its implicit dependence on how pre-activations are computed. The latter is inherited from model parameterisation. To mitigate the former, we exploited low dimensionality in upper convolutional layers of Bayesian CNNs and latent spaces of VAEs, thereby requiring less iterations of the conjugate gradient algorithm.

While our work has focused on the case where the matrix $\mathbf{V}$ is diagonal, we note that R2-G2 can be extended to the non-diagonal case, such as low-rank covariance matrices \citep{tomczak2020lowrank}, by defining appropriate matrix-vector product functions called within the conjugate gradient algorithm. Aside from likelihoods and ELBOs, we note that the R2-G2 estimator can be applied to other objective functions. A potential extension of our work can be to augment existing gradient estimators for variational inference in the \emph{multi-sample} setting \citep{roeder2017stl,rainforth2018tighter,tucker2018dreg,bauer2021gdreg} with the R2-G2 estimator and evaluate its effectiveness. We leave these directions for future work.

\section{Acknowledgements}
\label{section:acknowledgements}
    We would like to thank Andriy Mnih for helpful discussions and comments on drafts of this paper. Kevin H. Lam gratefully acknowledges his PhD funding from Google DeepMind. Thang D. Bui acknowledges support from the National Computing Infrastructure (NCI) Australia. George Deligiannidis acknowledges support from the Engineering and Physical Sciences Research Council [grant number EP/Y018273/1]. Yee Whye Teh acknowledges support from the Ministry of Digital Development and Information (MDDI) under the Singapore Global AI Visiting Professorship Program (Award No. AIVP-2024-002).

\bibliography{main.bib}

\begin{thebibliography}{45}
\providecommand{\natexlab}[1]{#1}
\providecommand{\url}[1]{\texttt{#1}}
\expandafter\ifx\csname urlstyle\endcsname\relax
  \providecommand{\doi}[1]{doi: #1}\else
  \providecommand{\doi}{doi: \begingroup \urlstyle{rm}\Url}\fi

\bibitem[Bauer and Mnih(2021)]{bauer2021gdreg}
Matthias Bauer and Andriy Mnih.
\newblock Generalized doubly reparameterized gradient estimators.
\newblock In \emph{Proceedings of the 38th International Conference on Machine Learning}, volume 139 of \emph{Proceedings of Machine Learning Research}, pages 738--747. PMLR, 18--24 Jul 2021.

\bibitem[Blackwell(1947)]{blackwell1947conditional}
David Blackwell.
\newblock Conditional expectation and unbiased sequential estimation.
\newblock \emph{The Annals of Mathematical Statistics}, 18\penalty0 (1):\penalty0 105--110, 1947.

\bibitem[Burda et~al.(2016)Burda, Grosse, and Salakhutdinov]{burda2016iwae}
Yuri Burda, Roger~B. Grosse, and Ruslan Salakhutdinov.
\newblock Importance weighted autoencoders.
\newblock In \emph{International Conference on Learning Representations}, 2016.

\bibitem[Dong et~al.(2020)Dong, Mnih, and Tucker]{dong2020disarm}
Zhe Dong, Andriy Mnih, and George Tucker.
\newblock Disarm: An antithetic gradient estimator for binary latent variables.
\newblock In \emph{Advances in Neural Information Processing Systems}, volume~33, pages 18637--18647. Curran Associates, Inc., 2020.

\bibitem[Eaton(1983)]{eaton1983multivariate}
Morris~L. Eaton.
\newblock \emph{Multivariate statistics : a vector space approach}.
\newblock Wiley series in probability and mathematical statistics. Probability and mathematical statistics. Wiley, New York, 1983.
\newblock ISBN 0471027766.

\bibitem[Figurnov et~al.(2018)Figurnov, Mohamed, and Mnih]{figurnov2018implicit}
Mikhail Figurnov, Shakir Mohamed, and Andriy Mnih.
\newblock Implicit reparameterization gradients.
\newblock In \emph{Advances in Neural Information Processing Systems}, volume~31. Curran Associates, Inc., 2018.

\bibitem[Glasserman(2003)]{glasserman2003monte}
Paul Glasserman.
\newblock \emph{{M}onte {C}arlo methods in financial engineering}, volume~53.
\newblock Springer, 2003.

\bibitem[Glynn(1990)]{glynn1990reinforce}
Peter~W. Glynn.
\newblock Likelihood ratio gradient estimation for stochastic systems.
\newblock \emph{Communications of the ACM}, 33\penalty0 (10):\penalty0 75–84, October 1990.
\newblock ISSN 0001-0782.

\bibitem[Greensmith et~al.(2001)Greensmith, Bartlett, and Baxter]{greensmith2001variance}
Evan Greensmith, Peter Bartlett, and Jonathan Baxter.
\newblock Variance reduction techniques for gradient estimates in reinforcement learning.
\newblock In \emph{Advances in Neural Information Processing Systems}, volume~14. MIT Press, 2001.

\bibitem[Hayami(2018)]{hayami2018convergence}
Ken Hayami.
\newblock Convergence of the conjugate gradient method on singular systems.
\newblock \emph{arXiv preprint arXiv:1809.00793}, 2018.

\bibitem[Huang et~al.(2021)Huang, Chen, Tsirigotis, and Courville]{huang2021cpflows}
Chin-Wei Huang, Ricky T.~Q. Chen, Christos Tsirigotis, and Aaron Courville.
\newblock Convex potential flows: Universal probability distributions with optimal transport and convex optimization.
\newblock In \emph{International Conference on Learning Representations}, 2021.

\bibitem[Huijben et~al.(2023)Huijben, Kool, Paulus, and van Sloun]{huijben2023survey}
Iris A.~M. Huijben, Wouter Kool, Max~B. Paulus, and Ruud J.~G. van Sloun.
\newblock A review of the {G}umbel-max trick and its extensions for discrete stochasticity in machine learning.
\newblock \emph{IEEE Transactions on Pattern Analysis and Machine Intelligence}, 45\penalty0 (2):\penalty0 1353--1371, 2023.

\bibitem[James(1978)]{james1978inverse}
M.~James.
\newblock The generalised inverse.
\newblock \emph{The Mathematical Gazette}, 62\penalty0 (420):\penalty0 109--114, 1978.
\newblock ISSN 00255572, 20566328.

\bibitem[Jankowiak and Obermeyer(2018)]{jankowiak2018pathwise}
Martin Jankowiak and Fritz Obermeyer.
\newblock Pathwise derivatives beyond the reparameterization trick.
\newblock In \emph{Proceedings of the 35th International Conference on Machine Learning}, volume~80 of \emph{Proceedings of Machine Learning Research}, pages 2235--2244. PMLR, 10--15 Jul 2018.

\bibitem[Kaasschieter(1988)]{kaasschieter1988preconditioned}
Erik~F Kaasschieter.
\newblock Preconditioned conjugate gradients for solving singular systems.
\newblock \emph{Journal of Computational and Applied mathematics}, 24\penalty0 (1-2):\penalty0 265--275, 1988.

\bibitem[Kingma and Ba(2015)]{kingma2015adam}
Diederik~P. Kingma and Jimmy Ba.
\newblock Adam: {A} method for stochastic optimization.
\newblock In \emph{International Conference on Learning Representations}, 2015.

\bibitem[Kingma and Welling(2014)]{kingma2014vae}
Diederik~P. Kingma and Max Welling.
\newblock Auto-encoding variational {B}ayes.
\newblock In \emph{International Conference on Learning Representations}, 2014.

\bibitem[Kingma et~al.(2015)Kingma, Salimans, and Welling]{kingma2015local}
Durk~P Kingma, Tim Salimans, and Max Welling.
\newblock Variational dropout and the local reparameterization trick.
\newblock In \emph{Advances in Neural Information Processing Systems}, volume~28. Curran Associates, Inc., 2015.

\bibitem[Kool et~al.(2020)Kool, van Hoof, and Welling]{kool2020unordered}
Wouter Kool, Herke van Hoof, and Max Welling.
\newblock Estimating gradients for discrete random variables by sampling without replacement.
\newblock In \emph{International Conference on Learning Representations}, 2020.

\bibitem[Krizhevsky and Hinton(2009)]{krizhevsky2009learning}
Alex Krizhevsky and Geoffrey Hinton.
\newblock Learning multiple layers of features from tiny images.
\newblock Technical report, University of Toronto, Toronto, Ontario, 2009.

\bibitem[Lake et~al.(2015)Lake, Salakhutdinov, and Tenenbaum]{lake2015omniglot}
Brenden~M. Lake, Ruslan Salakhutdinov, and Joshua~B. Tenenbaum.
\newblock Human-level concept learning through probabilistic program induction.
\newblock \emph{Science}, 350\penalty0 (6266):\penalty0 1332--1338, 2015.

\bibitem[LeCun et~al.(2010)LeCun, Cortes, and Burges]{lecun2010mnist}
Yann LeCun, Corinna Cortes, and CJ~Burges.
\newblock {MNIST} handwritten digit database.
\newblock \emph{ATT Labs}, 2, 2010.

\bibitem[Liu et~al.(2019)Liu, Regier, Tripuraneni, Jordan, and Mcauliffe]{liu2019rbreinforce}
Runjing Liu, Jeffrey Regier, Nilesh Tripuraneni, Michael Jordan, and Jon Mcauliffe.
\newblock {R}ao-{B}lackwellized stochastic gradients for discrete distributions.
\newblock In \emph{Proceedings of the 36th International Conference on Machine Learning}, volume~97 of \emph{Proceedings of Machine Learning Research}, pages 4023--4031. PMLR, 09--15 Jun 2019.

\bibitem[Mohamed et~al.(2020)Mohamed, Rosca, Figurnov, and Mnih]{mohamed2020survey}
Shakir Mohamed, Mihaela Rosca, Michael Figurnov, and Andriy Mnih.
\newblock {M}onte {C}arlo gradient estimation in machine learning.
\newblock \emph{Journal of Machine Learning Research}, 21\penalty0 (132):\penalty0 1--62, 2020.

\bibitem[Nocedal and Wright(1999)]{nocedal1999numerical}
Jorge Nocedal and Stephen~J. Wright.
\newblock \emph{Numerical Optimization}.
\newblock Springer, 1999.
\newblock ISBN 978-0-387-98793-4.

\bibitem[Paszke et~al.(2019)Paszke, Gross, Massa, Lerer, Bradbury, Chanan, Killeen, Lin, Gimelshein, Antiga, Desmaison, Kopf, Yang, DeVito, Raison, Tejani, Chilamkurthy, Steiner, Fang, Bai, and Chintala]{pytorch2019}
Adam Paszke, Sam Gross, Francisco Massa, Adam Lerer, James Bradbury, Gregory Chanan, Trevor Killeen, Zeming Lin, Natalia Gimelshein, Luca Antiga, Alban Desmaison, Andreas Kopf, Edward Yang, Zachary DeVito, Martin Raison, Alykhan Tejani, Sasank Chilamkurthy, Benoit Steiner, Lu~Fang, Junjie Bai, and Soumith Chintala.
\newblock Py{T}orch: An imperative style, high-performance deep learning library.
\newblock In \emph{Advances in Neural Information Processing Systems}, volume~32. Curran Associates, Inc., 2019.

\bibitem[Paulus et~al.(2021)Paulus, Maddison, and Krause]{paulus2021gumbelrao}
Max~B. Paulus, Chris~J. Maddison, and Andreas Krause.
\newblock {R}ao-{B}lackwellizing the straight-through gumbel-softmax gradient estimator.
\newblock In \emph{International Conference on Learning Representations}, 2021.

\bibitem[Planitz(1979)]{planitz1979leastsquares}
M.~Planitz.
\newblock Inconsistent systems of linear equations.
\newblock \emph{The Mathematical Gazette}, 63\penalty0 (425):\penalty0 181--185, 1979.
\newblock ISSN 00255572, 20566328.

\bibitem[Price(1958)]{price1958gaussian}
Robert Price.
\newblock A useful theorem for nonlinear devices having {G}aussian inputs.
\newblock \emph{{IRE} Transactions on Information Theory}, 4\penalty0 (2):\penalty0 69--72, 1958.

\bibitem[Rainforth et~al.(2018)Rainforth, Kosiorek, Le, Maddison, Igl, Wood, and Teh]{rainforth2018tighter}
Tom Rainforth, Adam Kosiorek, Tuan~Anh Le, Chris Maddison, Maximilian Igl, Frank Wood, and Yee~Whye Teh.
\newblock Tighter variational bounds are not necessarily better.
\newblock In \emph{Proceedings of the 35th International Conference on Machine Learning}, volume~80 of \emph{Proceedings of Machine Learning Research}, pages 4277--4285. PMLR, 10--15 Jul 2018.

\bibitem[Rao et~al.(1992)]{rao1992information}
C~Radhakrishna Rao et~al.
\newblock Information and the accuracy attainable in the estimation of statistical parameters.
\newblock \emph{Breakthroughs in statistics}, pages 235--247, 1992.

\bibitem[Rezende et~al.(2014)Rezende, Mohamed, and Wierstra]{rezende2014stochastic}
Danilo~Jimenez Rezende, Shakir Mohamed, and Daan Wierstra.
\newblock Stochastic backpropagation and approximate inference in deep generative models.
\newblock In \emph{Proceedings of the 31st International Conference on Machine Learning}, volume~32 of \emph{Proceedings of Machine Learning Research}, pages 1278--1286, Bejing, China, 22--24 Jun 2014. PMLR.

\bibitem[Roeder et~al.(2017)Roeder, Wu, and Duvenaud]{roeder2017stl}
Geoffrey Roeder, Yuhuai Wu, and David~K Duvenaud.
\newblock Sticking the landing: Simple, lower-variance gradient estimators for variational inference.
\newblock In \emph{Advances in Neural Information Processing Systems}, volume~30. Curran Associates, Inc., 2017.

\bibitem[Salakhutdinov and Murray(2008)]{salakhutdinov2008dynamic}
Ruslan Salakhutdinov and Iain Murray.
\newblock On the quantitative analysis of deep belief networks.
\newblock In \emph{Machine Learning, Proceedings of the Twenty-Fifth International Conference {(ICML} 2008), Helsinki, Finland, June 5-9, 2008}, volume 307 of \emph{{ACM} International Conference Proceeding Series}, pages 872--879. {ACM}, 2008.

\bibitem[Salimans(2014)]{salimans2014implementing}
Tim Salimans.
\newblock Implementing and automating fixed-form variational posterior approximation through stochastic linear regression.
\newblock \emph{arXiv preprint arXiv:1401.2135}, 2014.

\bibitem[Salimans and Knowles(2013)]{salimans2013fixed}
Tim Salimans and David~A Knowles.
\newblock Fixed-form variational posterior approximation through stochastic linear regression.
\newblock \emph{Bayesian Analysis}, 8\penalty0 (4):\penalty0 837--882, 2013.

\bibitem[Salimans and Knowles(2014)]{salimans2014using}
Tim Salimans and David~A Knowles.
\newblock On using control variates with stochastic approximation for variational {B}ayes and its connection to stochastic linear regression.
\newblock \emph{arXiv preprint arXiv:1401.1022}, 2014.

\bibitem[Simonyan and Zisserman(2015)]{simonyan2015vgg}
Karen Simonyan and Andrew Zisserman.
\newblock Very deep convolutional networks for large-scale image recognition.
\newblock In \emph{International Conference on Learning Representations}, 2015.

\bibitem[Srivastava et~al.(2014)Srivastava, Hinton, Krizhevsky, Sutskever, and Salakhutdinov]{srivastava2014dropout}
Nitish Srivastava, Geoffrey Hinton, Alex Krizhevsky, Ilya Sutskever, and Ruslan Salakhutdinov.
\newblock Dropout: A simple way to prevent neural networks from overfitting.
\newblock \emph{Journal of Machine Learning Research}, 15\penalty0 (56):\penalty0 1929--1958, 2014.

\bibitem[Titsias and Lázaro-Gredilla(2014)]{titsias2014doubly}
Michalis Titsias and Miguel Lázaro-Gredilla.
\newblock Doubly stochastic variational {B}ayes for non-conjugate inference.
\newblock In \emph{Proceedings of the 31st International Conference on Machine Learning}, volume~32 of \emph{Proceedings of Machine Learning Research}, pages 1971--1979, Bejing, China, 22--24 Jun 2014. PMLR.

\bibitem[Tomczak et~al.(2020)Tomczak, Swaroop, and Turner]{tomczak2020lowrank}
Marcin Tomczak, Siddharth Swaroop, and Richard Turner.
\newblock Efficient low rank gaussian variational inference for neural networks.
\newblock In H.~Larochelle, M.~Ranzato, R.~Hadsell, M.F. Balcan, and H.~Lin, editors, \emph{Advances in Neural Information Processing Systems}, volume~33, pages 4610--4622. Curran Associates, Inc., 2020.

\bibitem[Tucker et~al.(2019)Tucker, Lawson, Gu, and Maddison]{tucker2018dreg}
George Tucker, Dieterich Lawson, Shixiang Gu, and Chris~J. Maddison.
\newblock Doubly reparameterized gradient estimators for {M}onte {C}arlo objectives.
\newblock In \emph{International Conference on Learning Representations}, 2019.

\bibitem[Williams(1992)]{williams1992reinforce}
Ronald~J. Williams.
\newblock Simple statistical gradient-following algorithms for connectionist reinforcement learning.
\newblock \emph{Machine Learning}, 8:\penalty0 229--256, 1992.

\bibitem[Xiao et~al.(2017)Xiao, Rasul, and Vollgraf]{xiao2017fmnist}
Han Xiao, Kashif Rasul, and Roland Vollgraf.
\newblock Fashion-{MNIST}: a novel image dataset for benchmarking machine learning algorithms.
\newblock \emph{CoRR}, abs/1708.07747, 2017.

\bibitem[Xu et~al.(2019)Xu, Quiroz, Kohn, and Sisson]{xu2019variance}
Ming Xu, Matias Quiroz, Robert Kohn, and Scott~A. Sisson.
\newblock Variance reduction properties of the reparameterization trick.
\newblock In \emph{Proceedings of the Twenty-Second International Conference on Artificial Intelligence and Statistics}, volume~89 of \emph{Proceedings of Machine Learning Research}, pages 2711--2720. PMLR, 16--18 Apr 2019.

\end{thebibliography}
\bibliographystyle{plainnat}

\newpage
\section*{NeurIPS Paper Checklist}

\begin{enumerate}

\item {\bf Claims}
    \item[] Question: Do the main claims made in the abstract and introduction accurately reflect the paper's contributions and scope?
    \item[] Answer: \answerYes{} 
    \item[] Justification:  Our main theoretical and experimental contributions are clearly stated in the abstract and demonstrated in the paper. They reflect the paper’s contributions and scope.

\item {\bf Limitations}
    \item[] Question: Does the paper discuss the limitations of the work performed by the authors?
    \item[] Answer: \answerYes{} 
    \item[] Justification: We discuss the limitations of our work in Section \ref{section:conclusion}.

\item {\bf Theory assumptions and proofs}
    \item[] Question: For each theoretical result, does the paper provide the full set of assumptions and a complete (and correct) proof?
    \item[] Answer: \answerYes{} 
    \item[] Justification: All proofs are provided in the technical appendices and appropriately referenced.

    \item {\bf Experimental result reproducibility}
    \item[] Question: Does the paper fully disclose all the information needed to reproduce the main experimental results of the paper to the extent that it affects the main claims and/or conclusions of the paper (regardless of whether the code and data are provided or not)?
    \item[] Answer: \answerYes{} 
    \item[] Justification: We provide full experiment details in Appendix \ref{appendix:experiment_details} and the code to run our experiments.

\item {\bf Open access to data and code}
    \item[] Question: Does the paper provide open access to the data and code, with sufficient instructions to faithfully reproduce the main experimental results, as described in supplemental material?
    \item[] Answer: \answerYes{} 
    \item[] Justification: We provide the code needed to run our experiments.

\item {\bf Experimental setting/details}
    \item[] Question: Does the paper specify all the training and test details (e.g., data splits, hyperparameters, how they were chosen, type of optimizer, etc.) necessary to understand the results?
    \item[] Answer: \answerYes{} 
    \item[] Justification: We provide full experiment details in Appendix \ref{appendix:experiment_details}.

\item {\bf Experiment statistical significance}
    \item[] Question: Does the paper report error bars suitably and correctly defined or other appropriate information about the statistical significance of the experiments?
    \item[] Answer: \answerYes{} 
    \item[] Justification: We report error bars in all tabulated results from our experiments.

\item {\bf Experiments compute resources}
    \item[] Question: For each experiment, does the paper provide sufficient information on the computer resources (type of compute workers, memory, time of execution) needed to reproduce the experiments?
    \item[] Answer: \answerYes{} 
    \item[] Justification: We provide full details of compute resources in Appendix \ref{appendix:experiment_details}.
    
\item {\bf Code of ethics}
    \item[] Question: Does the research conducted in the paper conform, in every respect, with the NeurIPS Code of Ethics \url{https://neurips.cc/public/EthicsGuidelines}?
    \item[] Answer: \answerYes{} 
    \item[] Justification:  After careful review of the NeurIPS Code of Ethics, it is clear that the research conducted in this paper conforms with the Code of Ethics in every respect.

\item {\bf Broader impacts}
    \item[] Question: Does the paper discuss both potential positive societal impacts and negative societal impacts of the work performed?
    \item[] Answer: \answerNA{} 
    \item[] Justification: This paper is mostly methodological. While it is possible for advances in machine learning to bring societal impacts, our paper remains general and not specific to any application so it is unlikely to bring about any immediate societal impact.
    
\item {\bf Safeguards}
    \item[] Question: Does the paper describe safeguards that have been put in place for responsible release of data or models that have a high risk for misuse (e.g., pretrained language models, image generators, or scraped datasets)?
    \item[] Answer: \answerNA{} 
    \item[] Justification: This paper poses no such risks.

\item {\bf Licenses for existing assets}
    \item[] Question: Are the creators or original owners of assets (e.g., code, data, models), used in the paper, properly credited and are the license and terms of use explicitly mentioned and properly respected?
    \item[] Answer: \answerYes{} 
    \item[] Justification: All datasets used in our experiments are open-source and have been properly referenced in the paper and Appendix \ref{appendix:licenses}.

\item {\bf New assets}
    \item[] Question: Are new assets introduced in the paper well documented and is the documentation provided alongside the assets?
    \item[] Answer: \answerNA{} 
    \item[] Justification: This paper does not introduce new assets.

\item {\bf Crowdsourcing and research with human subjects}
    \item[] Question: For crowdsourcing experiments and research with human subjects, does the paper include the full text of instructions given to participants and screenshots, if applicable, as well as details about compensation (if any)? 
    \item[] Answer: \answerNA{} 
    \item[] Justification: This paper does not involve crowdsourcing nor research with human subjects.

\item {\bf Institutional review board (IRB) approvals or equivalent for research with human subjects}
    \item[] Question: Does the paper describe potential risks incurred by study participants, whether such risks were disclosed to the subjects, and whether Institutional Review Board (IRB) approvals (or an equivalent approval/review based on the requirements of your country or institution) were obtained?
    \item[] Answer: \answerNA{} 
    \item[] Justification: This paper does not involve crowdsourcing nor research with human subjects.

\item {\bf Declaration of LLM usage}
    \item[] Question: Does the paper describe the usage of LLMs if it is an important, original, or non-standard component of the core methods in this research? Note that if the LLM is used only for writing, editing, or formatting purposes and does not impact the core methodology, scientific rigorousness, or originality of the research, declaration is not required.
    \item[] Answer: \answerNA{} 
    \item[] Justification: This paper does not involve usage of LLMs.

\end{enumerate}

\newpage
\appendix

\section{Supporting results}
This section lists the results supporting the design of the R2-G2 estimator, namely the analytical form of conditional Gaussian distributions and the conjugate gradient algorithm.
\begin{lemma}[{\citep[Proposition 3.13]{eaton1983multivariate}}]
\label{lemma:mvn_properties}
Suppose we have a random vector $\pmb{\epsilon} \sim \mathcal{N}(\mathbf{0}_n, \mathbf{I}_n)$ where $\pmb{\epsilon} \in \R^n$ and a linear map defined as
\begin{align*}
    \mathbf{A}: \R^n \to \R^m, \quad \mathbf{A}(\mathbf{x}) = \mathbf{A}\mathbf{x}
\end{align*}
where $\mathbf{A} \in \R^{m \times n}$. Then we have the joint distribution
\begin{align*}
    \begin{bmatrix}
    \pmb{\epsilon}\\
    \mathbf{A}(\pmb{\epsilon})
    \end{bmatrix}
    \sim \mathcal{N}\left(
    \begin{bmatrix}
    \mathbf{0}_n\\
    \mathbf{0}_m
    \end{bmatrix},
    \begin{bmatrix}
    \mathbf{I}_n & \mathbf{A}^{\top}\\
    \mathbf{A} & \mathbf{A}\mathbf{A}^{\top}
    \end{bmatrix}
    \right),
\end{align*}
and the conditional distribution
\begin{align*}
    \pmb{\epsilon} | \mathbf{A}(\pmb{\epsilon}) = \mathbf{z}
    \sim \mathcal{N}\left(
    \mathbf{A}^{\top}\left(\mathbf{A}\mathbf{A}^{\top}\right)^{\dagger}\mathbf{z},
    \mathbf{I}_n - \mathbf{A}^{\top}\left(\mathbf{A}\mathbf{A}^{\top}\right)^{\dagger}\mathbf{A}
    \right)
\end{align*}
where $\left(\mathbf{A}\mathbf{A}^{\top}\right)^{\dagger}$ is the Moore-Penrose pseudo-inverse of $\mathbf{A}\mathbf{A}^{\top}$.
\end{lemma}

\begin{algorithm}
   \caption{Conjugate Gradient Algorithm}
   \label{algorithm:conjugate_gradient}
\begin{algorithmic}
   \STATE {\bfseries Input:} number of iterations $T$, matrix $\mathbf{A}\in \R^{m \times n}$, column vector $\mathbf{z} \in \R^m$.
   \STATE Initialise $\mathbf{x}_0 \in \R^m$.
   \STATE Set $\mathbf{r}_0 = \mathbf{A} \mathbf{A}^{\top}\mathbf{x}_0 - \mathbf{z}$.
   \STATE Set $\mathbf{p}_0 = -\mathbf{r}_0$.
   \STATE Set $k=0$.
   \WHILE{$k < T$ and $\mathbf{r}_k \neq \mathbf{0}$}
   \STATE Compute $\alpha_k = \frac{\mathbf{r}_k^{\top} \mathbf{r}_k}{\mathbf{p}_k^{\top} \mathbf{A} \mathbf{A}^{\top} \mathbf{p}_k}$.
   \STATE Compute $\mathbf{x}_{k+1} = \mathbf{x}_k + \alpha_k \mathbf{p}_k$.
   \STATE Compute $\mathbf{r}_{k+1} = \mathbf{r}_k + \alpha_k \mathbf{A} \mathbf{A}^{\top} \mathbf{p}_k$.
   \STATE Compute $\beta_{k+1} = \frac{\mathbf{r}_{k+1}^{\top} \mathbf{r}_{k+1}}{\mathbf{r}_k^{\top} \mathbf{r}_k}$.
   \STATE Compute $\mathbf{p}_{k+1} = -\mathbf{r}_{k+1} + \beta_{k+1} \mathbf{p}_k$.
   \STATE Increment $k$ by 1.
   \ENDWHILE
   \STATE {\bfseries Output:} $\mathbf{x}_{T}$.
\end{algorithmic}
\end{algorithm}

\newpage
\section{Proof for Proposition \ref{proposition:r2g2_key_properties}}
\label{appendix:r2g2_key_properties}
We first prove that the R2-G2 estimator is an unbiased estimator of $\nabla_{\pmb{\theta}}\mathbb{E}_{q_{\pmb{\theta}}}[\ell]$. Recall that 
\begin{align*}
    \widehat{\nabla_{\pmb{\theta}}\ell}^{R2\text{-}G2} &= \mathbb{E}_{\Tilde{q}_0}\left[\widehat{\nabla_{\pmb{\theta}}\ell}^{RT}\right].
\end{align*}
We note that $\widehat{\nabla_{\pmb{\theta}}\ell}^{R2\text{-}G2}$ is a conditional expectation with $\mathbf{W} \cdot g = \mathbf{z}$ as the conditioning variable. Then we have 
\begin{align*}
    \mathbb{E}_{q_{\mathbf{z}}}\left[\widehat{\nabla_{\pmb{\theta}}\ell}^{R2\text{-}G2}\right] &= \mathbb{E}_{q_{\mathbf{z}}}\left[\mathbb{E}_{\Tilde{q}_0}\left[\widehat{\nabla_{\pmb{\theta}}\ell}^{RT}\right]\right]\\
    &= \mathbb{E}_{q_0}\left[\widehat{\nabla_{\pmb{\theta}}\ell}^{RT}\right]\\
    &= \nabla_{\pmb{\theta}}\mathbb{E}_{q_{\pmb{\theta}}}[\ell]
\end{align*}
where the second equality follows from the law of iterated expectations. Since the reparameterisation gradient estimator $\widehat{\nabla_{\pmb{\theta}}\ell}^{RT}$ is an unbiased estimator of $\nabla_{\pmb{\theta}}\mathbb{E}_{q_{\pmb{\theta}}}[\ell]$, it follows that the $R2\text{-}G2$ estimator $\widehat{\nabla_{\pmb{\theta}}\ell}^{R2\text{-}G2}$ is also an unbiased estimator of $\nabla_{\pmb{\theta}}\mathbb{E}_{q_{\pmb{\theta}}}[\ell]$.

We are left to show that the R2-G2 estimator has less variance than the reparameterisation gradient estimator. We can write
\begin{align*}
    \mathbb{E}_{q_{\mathbf{z}}}\left[\left\|\widehat{\nabla_{\pmb{\theta}} \ell}^{R2\text{-}G2} -  \nabla_{\pmb{\theta}}\mathbb{E}_{q_{\pmb{\theta}}}[\ell]\right\|^2\right] &= \mathbb{E}_{q_{\mathbf{z}}}\left[\left\|\mathbb{E}_{\Tilde{q}_0}\left[\widehat{\nabla_{\pmb{\theta}}\ell}^{RT}\right] -  \nabla_{\pmb{\theta}}\mathbb{E}_{q_{\pmb{\theta}}}[\ell]\right\|^2\right]\\
    &= \mathbb{E}_{q_{\mathbf{z}}}\left[\left\|\mathbb{E}_{\Tilde{q}_0}\left[\widehat{\nabla_{\pmb{\theta}}\ell}^{RT} -  \nabla_{\pmb{\theta}}\mathbb{E}_{q_{\pmb{\theta}}}[\ell] \right]\right\|^2\right]\\
    &\leq \mathbb{E}_{q_{\mathbf{z}}}\left[\mathbb{E}_{\Tilde{q}_0}\left[\left\|\widehat{\nabla_{\pmb{\theta}}\ell}^{RT} -  \nabla_{\pmb{\theta}}\mathbb{E}_{q_{\pmb{\theta}}}[\ell]\right\|^2 \right]\right]\\
    &= \mathbb{E}_{q_0}\left[\left\|\widehat{\nabla_{\pmb{\theta}}\ell}^{RT} -  \nabla_{\pmb{\theta}}\mathbb{E}_{q_{\pmb{\theta}}}[\ell]\right\|^2 \right]
\end{align*}
where the inequality results from using Jensen's inequality, and the last equality comes from the law of iterated expectations.

\section{Proof of Theorem \ref{theorem:local_reparam_as_r2g2}}
\label{appendix:local_reparam_as_r2g2}
Suppose we have a BNN linear layer where weights are independent Gaussian random variables. Given an input $\mathbf{x} \in \R^n$, the pre-activations and parameters of these linear layers are respectively given by
\begin{align*}
    \mathbf{z} =
    \begin{bmatrix}
    \mathbf{x}^{\top} g(\pmb{\epsilon}^{(1)}, \pmb{\theta}^{(1)}) \\
    \vdots \\
    \mathbf{x}^{\top} g(\pmb{\epsilon}^{(m)}, \pmb{\theta}^{(m)})\\
    \end{bmatrix}
    \in \R^{m}
\end{align*}
where $\pmb{\theta}^{(i)}= \{\pmb{\mu}^{(i)}, \pmb{\tau}^{(i)}\}$ for $i=1,\dots,m$. We proceed by considering a fixed $i$. 

By setting $\mathbf{V}^{(i)}=(\pmb{\Sigma}^{(i)})^{\frac{1}{2}}=\left(\texttt{diag}(\pmb{\tau}^{(i)})\right)^{\frac{1}{2}}$ and $\mathbf{W} = \mathbf{x}^{\top}$ within Equation \ref{equation:reparam_gradient_estimator_decomposed}, the reparameterisation gradient estimator is given by
\begin{align*}
    \widehat{\nabla_{\pmb{\theta}^{(i)}}[\ell]}^{RT}
    &=
    \left(
    \frac{\partial \Tilde{\ell}}{\partial z_i} \cdot \mathbf{x}^{\top} \cdot 
    \begin{bmatrix}
    \begin{array}{c|c}
    \mathbf{I}_n & \frac{1}{2}(\pmb{\Sigma}^{(i)})^{-\frac{1}{2}} \odot (\mathbf{1}_n(\pmb{\epsilon}^{(i)})^{\top}) 
    \end{array}
    \end{bmatrix}
    \right)^{\top}.
\end{align*}
We proceed with the idea of Rao-Blackwellisation from the R2-G2 estimator by conditioning on the pre-activation $z_i = \mathbf{x}^{\top} \cdot g(\pmb{\epsilon}^{(i)}, \pmb{\theta}^{(i)}) \in \mathbf{z}$. This amounts to setting $\mathbf{A}=\mathbf{x}^{\top}\mathbf{V}^{(i)}$ in Definition \ref{definition:r2g2}.

We note that $\mathbf{x}^{\top} \cdot g(\pmb{\epsilon}^{(i)}, \pmb{\theta}^{(i)}) = z_i$ is equivalent to $\mathbf{A} \pmb{\epsilon}^{(i)} = z_i - \mathbf{x}^{\top}\pmb{\mu}^{(i)} = \Tilde{z}_i$. Denote $\Tilde{\pmb{\epsilon}}^{(i)}$ as $\pmb{\epsilon}^{(i)} | \mathbf{A} \pmb{\epsilon}^{(i)}  = \Tilde{z}_i$ with distribution denoted $\Tilde{q}_0^{(i)}$. Using Lemma \ref{lemma:mvn_properties}, we can write the distribution $\Tilde{\pmb{\epsilon}}^{(i)} \sim \Tilde{q}_0^{(i)}$ as
\begin{align*}
    \Tilde{q}_0^{(i)} =\mathcal{N}\left((\pmb{\Sigma}^{(i)})^{\frac{1}{2}}\mathbf{x}\left(\frac{\Tilde{z}_i}{\sum_{j=1}^{n}x_{j}^2 \left(\sigma_j^{(i)}\right)^2}\right), \mathbf{I}_{n} - (\pmb{\Sigma}^{(i)})^{\frac{1}{2}}\mathbf{x}\left(\sum_{j=1}^{n}x_{j}^2 \left(\sigma_j^{(i)}\right)^2\right)^{-1}\mathbf{x}^{\top}(\pmb{\Sigma}^{(i)})^{\frac{1}{2}}\right).
\end{align*}

Define the transformation of a vector to a diagonal matrix as 
\begin{align*}
    D: \R^n \to \R^{n \times n}, \quad D(\mathbf{x}) = \sum_{k=1}^{n} e_k^{\top}\mathbf{x} e_k e_k^{\top}
\end{align*}
where $e_k \in \R^n$ is the $k$-th standard basis vector. Using the linearity of expectations, we have
\begin{align*}
    \frac{1}{2}\mathbf{x}^{\top} \cdot \mathbb{E}_{\Tilde{q}_0^{(i)}}\left[(\pmb{\Sigma}^{(i)})^{-\frac{1}{2}} \odot  \mathbf{1}_n(\pmb{\epsilon}^{(i)})^{\top}\right] &= \frac{1}{2}\mathbf{x}^{\top} \cdot \left[(\pmb{\Sigma}^{(i)})^{-\frac{1}{2}} \odot  \mathbf{1}_n\left(\mathbb{E}_{\Tilde{q}_0^{(i)}}[\pmb{\epsilon}^{(i)}]\right)^{\top}\right]\\
    &= \frac{1}{2} \left(\frac{\Tilde{z}_i}{\sum_{j=1}^{n}x_{j}^2 \left(\sigma_j^{(i)}\right)^2}\right) \mathbf{x}^{\top} \cdot \left[(\pmb{\Sigma}^{(i)})^{-\frac{1}{2}} \odot  (\mathbf{1}_n \mathbf{x}^{\top} (\pmb{\Sigma}^{(i)})^{\frac{1}{2}})\right]\\
    &= \frac{1}{2} \left(\frac{\Tilde{z}_i}{\sum_{j=1}^{n}x_{j}^2 \left(\sigma_j^{(i)}\right)^2}\right) \mathbf{x}^{\top}D\left(\mathbf{x}\right)\\
    &= \frac{1}{2} \left(\frac{z_i- \sum_{j=1}^{n}x_{j}\mu_j^{(i)}}{\sum_{j=1}^{n}x_{j}^2 \left(\sigma_j^{(i)}\right)^2}\right) \left(\mathbf{x} \odot \mathbf{x}\right)^{\top}\\
    &\overset{d}{=} \frac{1}{2}\left(\sum_{j=1}^{n}x_{j}^2\left(\sigma_j^{(i)}\right)^2\right)^{-\frac{1}{2}} \xi_i \left(\mathbf{x} \odot \mathbf{x}\right)^{\top}
\end{align*}
where $\mathbf{x}^{\top}D\left(\mathbf{x}\right) = \left(\mathbf{x} \odot \mathbf{x}\right)^{\top}$ is a row vector with entries $\left\{x_{j}^2\right\}_{j=1}^{n}$ and applying the reparameterisation trick gives us 
\begin{align*}
    z_i \overset{d}{=} \sum_{j=1}^{n}x_{j}\mu_j^{(i)} + \left(\sum_{i=1}^{n}x_{j}^2 \left(\sigma_j^{(i)}\right)^2\right)^{\frac{1}{2}} \xi_i \sim \mathcal{N}\left(\sum_{j=1}^{n}x_{j}\mu_j^{(i)}, \sum_{j=1}^{n}x_{j}^2 \left(\sigma_j^{(i)}\right)^2\right)
\end{align*}
for $\xi_i \sim \mathcal{N}(0,1)$. Since we also have $\mathbf{x}^{\top} \cdot \mathbb{E}_{\Tilde{q}_0^{(i)}}\left[\mathbf{I}_n\right] = \mathbf{x}^{\top}$, the Jacobian of $z_i = \mathbf{x}^{\top} \cdot g(\pmb{\epsilon}^{(i)}, \pmb{\theta}^{(i)})$ is
\begin{align*}
    J_{z_i}(\pmb{\theta}^{(i)}) &=\mathbf{x}^{\top} \cdot \mathbb{E}_{\Tilde{q}_0^{(i)}}[J_{g^{(i)}}] \\
    &\overset{d}{=} 
    \begin{bmatrix}
    \begin{array}{c|c}
         \mathbf{x}^{\top} & \frac{1}{2}\left(\sum_{j=1}^{n}x_{j}^2\left(\sigma_j^{(i)}\right)^2\right)^{-\frac{1}{2}} \xi_i \left(\mathbf{x} \odot \mathbf{x}\right)^{\top}
    \end{array}
    \end{bmatrix}.
\end{align*}

Hence, we have
\begin{align*}
    \widehat{\nabla_{\pmb{\theta}^{(i)}}\ell}^{R2\text{-}G2} \overset{d}{=} \widehat{\nabla_{\pmb{\theta}^{(i)}}\ell}^{LRT}.
\end{align*}
This shows that the local reparameterisation gradient estimator is equivalent in distribution to a Rao-Blackwellised reparameterisation gradient estimator. 

We are left to show that the local reparameterisation estimator has lower variance than the global reparameterisation estimator. We have
\begin{align*}
    \mathbb{E}_{\Tilde{q}_{z_i}}\left[\left\| \widehat{\nabla_{\pmb{\theta}^{(i)}}\ell}^{LRT} -  \nabla_{\pmb{\theta}^{(i)}}\mathbb{E}_{q_{\pmb{\theta}^{(i)}}}[\ell]\right\|^2\right] &= \mathbb{E}_{\Tilde{q}_{z_i}}\left[\left\|\widehat{\nabla_{\pmb{\theta}^{(i)}}\ell}^{R2\text{-}G2} -  \nabla_{\pmb{\theta}^{(i)}}\mathbb{E}_{q_{\pmb{\theta}^{(i)}}}[\ell]\right\|^2\right]\\
    &= \mathbb{E}_{\Tilde{q}_{z_i}}\left[\left\|\mathbb{E}_{\Tilde{q}_0}\left[\widehat{\nabla_{\pmb{\theta}^{(i)}}\ell}^{RT}\right] -  \nabla_{\pmb{\theta}^{(i)}}\mathbb{E}_{q_{\pmb{\theta}^{(i)}}}[\ell]\right\|^2\right]\\
    &= \mathbb{E}_{\Tilde{q}_{z_i}}\left[\left\|\mathbb{E}_{\Tilde{q}_0}\left[\widehat{\nabla_{\pmb{\theta}^{(i)}}\ell}^{RT} -  \nabla_{\pmb{\theta}^{(i)}}\mathbb{E}_{q_{\pmb{\theta}^{(i)}}}[\ell]\right]\right\|^2\right]\\
    &\leq\mathbb{E}_{\Tilde{q}_{z_i}}\left[\mathbb{E}_{\Tilde{q}_0}\left[\left\|\widehat{\nabla_{\pmb{\theta}^{(i)}}\ell}^{RT} -  \nabla_{\pmb{\theta}^{(i)}}\mathbb{E}_{q_{\pmb{\theta}^{(i)}}}[\ell]\right\|^2\right]\right]\\
    &= \mathbb{E}_{q_0^{(i)}}\left[\left\|\widehat{\nabla_{\pmb{\theta}^{(i)}}\ell}^{RT} -  \nabla_{\pmb{\theta}^{(i)}}\mathbb{E}_{q_{\pmb{\theta}^{(i)}}}[\ell]\right\|^2 \right]
\end{align*}
where the inequality results from using Jensen's inequality and the last equality comes from the law of iterated expectations.

\section{Computation of Conditional Mean}
\label{appendix:verifying_conditional_mean}
Recall that the image and kernel of a linear map $\mathbf{T}:\R^n \to \R^m$ are defined respectively as $\texttt{im}(\mathbf{T}) = \{\mathbf{T}\mathbf{v}: \mathbf{v} \in \R^n\}$ and $\texttt{ker}(\mathbf{T}) = \{\mathbf{v}: \mathbf{T}\mathbf{v}=\mathbf{0}\}$. Since $\mathbf{A} \pmb{\epsilon} \in \texttt{im}(\mathbf{A})=\texttt{im}(\mathbf{A}\mathbf{A}^{\top})$, we note that Equation \ref{equation:normal_equation} has solutions in the form 
\begin{align*}
    \left(\mathbf{A}\mathbf{A}^{\top}\right)^{\dagger}\mathbf{A} \pmb{\epsilon} + (\mathbf{I}_m - \left(\mathbf{A}\mathbf{A}^{\top}\right)^{\dagger}\mathbf{A}\mathbf{A}^{\top})\mathbf{y}
\end{align*}
for any $\mathbf{y} \in \R^m$ (see \cite{james1978inverse,planitz1979leastsquares}). Here, $\left(\mathbf{A}\mathbf{A}^{\top}\right)^{\dagger}\mathbf{A} \pmb{\epsilon} \in \texttt{im}(\mathbf{A}\mathbf{A}^{\top}) = \texttt{im}(\mathbf{A})$ and $(\mathbf{I}_m - \left(\mathbf{A}\mathbf{A}^{\top}\right)^{\dagger}\mathbf{A}\mathbf{A}^{\top})\mathbf{y} \in \texttt{ker}(\mathbf{A}\mathbf{A}^{\top}) = \texttt{ker}(\mathbf{A}^{\top})$. That is, we can recast the matrix-vector product $\left(\mathbf{A}\mathbf{A}^{\top}\right)^{\dagger}\mathbf{A} \pmb{\epsilon}$ as a solution to Equation \ref{equation:normal_equation}, up to an additive term from $\texttt{ker}(\mathbf{A}^{\top})$. For any solution $\pmb{\beta}^{*}$ to Equation \ref{equation:normal_equation}, it then follows that 
\begin{align*}
    \mathbf{A}^{\top}\pmb{\beta}^{*} = \mathbf{A}^{\top}\left(\mathbf{A}\mathbf{A}^{\top}\right)^{\dagger}\mathbf{A} \pmb{\epsilon} = \pmb{\epsilon}^{*}.
\end{align*}

\newpage
\section{Experiment details}
\label{appendix:experiment_details}
\paragraph{Compute resources} All experiments were run on a single NVIDIA V100 GPU.

\paragraph{Architectures} Model architectures for BNN and VAE experiments can be found in Section \ref{section:experiments}. We excluded batch normalisation and dropout layers from model architectures.

\paragraph{Optimisation} We used a batch size of 80 and the Adam optimiser for all experiments \cite{kingma2015adam}. We do not add regularisation such as weight decay, dropout or batch normalisation layers. Other optimisation parameters are listed in Table \ref{table:hyperparameters_running_times}.

For consistency with work on gradient estimators, we report the number of optimisation/SGD steps. For BNN experiments on the MNIST and CIFAR-10 datasets, this is equivalent to training with a batch size of $80$ for $20$ epochs.

\begin{table}[t]
\caption{Optimisation hyperparameters and running times for experiments.}
\label{table:hyperparameters_running_times}
\begin{center}
\begin{tabular}{lcccccc}
\toprule
Model & Estimator & Steps & Learning Rate & Steps/s & Run Time Limit (hours)\\
\midrule
Bayesian MLP & R2-G2  & 15,000 & 0.0001 & 13.57 & 0.5\\
Bayesian MLP & LRT & 15,000 & 0.0001 & 48.25 & 0.5\\
Bayesian MLP & RT & 15,000 & 0.0001 & 10.52 & 0.5\\
Bayesian CNN & R2-G2 & 12,500 & 0.0001 & 0.68 & 10\\
Bayesian CNN & RT & 12,500 & 0.0001 & 2.31 & 5\\
One-layer VAE & R2-G2 & 100,000 & 0.0003 & 47.96 & 1\\
One-layer VAE & RT & 100,000 & 0.0003 & 127.66 & 1\\
Two-layer VAE & R2-G2 & 100,000 & 0.0003 & 20.22 & 1.5\\
Two-layer VAE & RT & 100,000 & 0.0003 & 78.72 & 1.5\\
Three-layer VAE & R2-G2 & 100,000 & 0.0003 & 17.11 & 2.5\\
Three-layer VAE & RT & 100,000 & 0.0003 & 48.41 & 2.5\\
\bottomrule
\end{tabular}
\end{center}
\end{table}

\section{Comparison of Computational Cost of Gradient Estimators}
\label{appendix:comparison_computational_cost}
In general, the worst-case complexity of computational costs from the conjugate gradient (CG) algorithm is $\mathcal{O}(m^3)$ when one is attempting to invert a $m \times m$ dense matrix $\mathbf{A}$ without knowing its structure. Here the worst-case would be running $m$ iterations of CG with each iteration requiring $m^2$ flops for a matrix-vector product.

In the mean-field setting, which is the main focus of our work, we know additional structure about the matrix $\mathbf{A}$. Specifically, it can factorise as $\mathbf{A} = \mathbf{W}\mathbf{V}\mathbf{V}^{\top}\mathbf{W}^{\top}$ where $\mathbf{V} \in \mathbb{R}^{n \times n}$ is diagonal and $\mathbf{W} \in \mathbb{R}^{m \times n}$. Moreover, we know the rank of $\mathbf{V}$ and the maximum rank of $\mathbf{W}$, so CG only needs to be run for $k=\min(m,n)$ iterations at most, by using the property of ranks that $\text{rank}(AB) \leq min(\text{rank}(A), \text{rank}(B))$ for matrices $A,B$.

A forward pass for a linear layer using the global reparameterisation trick would use a total of $(2m + 1)n$ flops for matrix-vector products. Using the R2-G2 estimator runs at most $k$ iterations of the CG algorithm with each iteration requiring $(2m + 1)n$ flops for matrix-vector products, so an additional $(2m + 1)nk$ flops would be incurred at most. 

For BNNs, we would have $n > m^2$ since there are more weights than pre-activations, so the cost complexity would be $\mathcal{O}(n)$ for a forward pass using the global reparameterisation trick and an additional $\mathcal{O}(m^2n)$ for R2-G2. For VAEs, we would have $m > n$ since we map low-dimensional latents to a higher-dimensional space, so the cost complexity would be $\mathcal{O}(m)$ for a forward pass using the global reparameterisation trick and an additional $\mathcal{O}(mn^2)$ for R2-G2.

\newpage
\section{Additional plots of gradient variance}
\begin{figure*}[ht]
  \centering
  \begin{tikzpicture}
  \begin{axis}[
      width=0.8*\textwidth,
      height=0.6*0.8*\textwidth,
      grid=major, 
      grid style={thick,gray!30},
      xlabel= Epoch, 
      ylabel= Log Gradient Variance,
      legend pos= outer north east,
      x tick label style={ultra thick, anchor=north},
      xmin=1, xmax=20,
      ymin=-4, ymax=-1,
      xtick={5, 10, 15, 20},
      ytick={-4, -3.5, -3, -2.5, -2, -1.5},
      yticklabels={-4, -3.5, -3, -2.5, -2, -1.5},
      scaled x ticks = false, 
      x tick label style={/pgf/number format/.cd, fixed, fixed zerofill, int detect, 1000 sep={}}
    ]
    \addplot [ultra thick, dotted, blue, mark=*,text mark as node=true] 
    table[x=epoch,y=log_gradient_variance,col sep=comma, discard if not={estimator}{RT}, discard if not={layer}{bottom}]{results/bnn_mnist_grad_var.csv};
    \addplot [ultra thick, blue, mark=*,text mark as node=true] table[x=epoch,y=log_gradient_variance,col sep=comma, discard if not={estimator}{LRT}, discard if not={layer}{bottom}]{results/bnn_mnist_grad_var.csv};
    \addplot [ultra thick, dash pattern=on 6pt off 2pt, orange, mark=+,text mark as node=true]
    table[x=epoch,y=log_gradient_variance,col sep=comma, discard if not={estimator}{R2-G2}, discard if not={layer}{bottom}]{results/bnn_mnist_grad_var.csv};

    \legend{
        RT,
        LRT,
        R2-G2,
    }
  \end{axis}
\end{tikzpicture}
  \caption{Log gradient variance v.s. epoch for the bottom layer of a Bayesian MLP trained on \emph{MNIST} over 5 runs. We compare the variance of gradients when training using the reparameterisation (RT), local reparameterisation (LRT) and R2-G2 estimators.}
  \label{figure:bnn_mnist_grad_var_bottom}
\end{figure*}

\begin{figure*}
  \centering
  \begin{tikzpicture}
  \begin{axis}[
      width=0.8*\textwidth,
      height=0.6*0.8*\textwidth,
      grid=major, 
      grid style={thick,gray!30},
      xlabel= Epoch, 
      ylabel= Log Gradient Variance,
      legend pos= outer north east,
      x tick label style={ultra thick, anchor=north},
      xmin=10, xmax=40,
      ymin=-3.5, ymax=-1,
      xtick={10, 20, 30, 40},
      ytick={-3, -2.5, -2, -1.5},
      yticklabels={-3, -2.5, -2, -1.5},
      scaled x ticks = false, 
      x tick label style={/pgf/number format/.cd, fixed, fixed zerofill, int detect, 1000 sep={}}
    ]
    \addplot [ultra thick, dotted, blue, mark=*,text mark as node=true] 
    table[x=epoch,y=log_gradient_variance,col sep=comma, discard if not={estimator}{RT}, discard if not={layer}{top}]{results/bnn_cifar_grad_var.csv};
    \addplot [ultra thick, dash pattern=on 6pt off 2pt, orange, mark=+,text mark as node=true] table[x=epoch,y=log_gradient_variance,col sep=comma, discard if not={estimator}{R2-G2}, discard if not={layer}{top}]{results/bnn_cifar_grad_var.csv};
    
    \legend{
        RT,
        R2-G2,
    }
  \end{axis}
\end{tikzpicture}
  \caption{Log gradient variance v.s. epoch for the $8$-th convolutional layer of a Bayesian CNN trained on \emph{CIFAR-10} over 5 runs. We compare the variance of gradients when training using the reparameterisation (RT) and R2-G2 estimators.}
  \label{figure:bnn_cifar_grad_var_top}
\end{figure*}

\begin{figure*}
  \centering
  \begin{tikzpicture}
  \begin{axis}[
      width=0.8*\textwidth,
      height=0.6*0.8*\textwidth,
      grid=major, 
      grid style={thick,gray!30},
      xlabel= Epoch, 
      ylabel= Log Gradient Variance,
      legend pos= outer north east,
      x tick label style={ultra thick, anchor=north},
      xmin=10, xmax=40,
      ymin=-2.5, ymax=0,
      xtick={10, 20, 30, 40},
      ytick={-2.5, -2, -1.5, -1, -0.5, 0},
      yticklabels={-2.5, -2, -1.5, -1, -0.5, 0},
      scaled x ticks = false, 
      x tick label style={/pgf/number format/.cd, fixed, fixed zerofill, int detect, 1000 sep={}}
    ]
    \addplot [ultra thick, dotted, blue, mark=*,text mark as node=true] 
    table[x=epoch,y=log_gradient_variance,col sep=comma, discard if not={estimator}{RT}, discard if not={layer}{bottom}]{results/bnn_cifar_grad_var.csv};
    \addplot [ultra thick, dash pattern=on 6pt off 2pt, orange, mark=+,text mark as node=true] table[x=epoch,y=log_gradient_variance,col sep=comma, discard if not={estimator}{R2-G2}, discard if not={layer}{bottom}]{results/bnn_cifar_grad_var.csv};
    
    \legend{
        RT,
        R2-G2,
    }
  \end{axis}
\end{tikzpicture}
  \caption{Log gradient variance v.s. epoch for the $5$-th convolutional layer of a Bayesian CNN trained on \emph{CIFAR-10} over 5 runs. We compare the variance of gradients when training using the reparameterisation (RT) and R2-G2 estimators.}
  \label{figure:bnn_cifar_grad_var_bottom}
\end{figure*}

\newpage
\section{Experiments on single-layer Variational Autoencoders}
\label{appendix:1vae_experiments}
\begin{table}[h]
\caption{Test variational lower bounds for VAEs using the R2-G2 and Reparameterisation (RT) estimators. Higher is better. Error bars denote $\pm 1.96$ standard errors $(\sigma / \sqrt{5})$ over 5 runs.}
\label{table:vae_results}
\begin{center}
\begin{tabular}{lcccc}
\toprule
\# VAE Layers & Estimator & MNIST & Omniglot & Fashion-MNIST \\ 
\midrule
1 & R2-G2 & $-94.39 \pm 0.42$ & $\mathbf{-117.61 \pm 2.12}$ & $-238.65 \pm 0.26$\\
& RT & $\mathbf{-94.22 \pm 0.24}$ & $-117.64 \pm 2.12$ & $\mathbf{-238.64 \pm 0.25}$\\
\bottomrule
\end{tabular}
\end{center}
\end{table}

The R2-G2 estimator yielded limited gains on performance for single-layer VAEs. In this setting, the R2-G2 estimator only impacts the optimisation of the encoder. This motivates our experiments for hierarchical VAEs where we only apply the R2-G2 estimator within the decoder (i.e. $\mathbf{W}$ and $\mathbf{V}$ are both matrix parameters in the decoder).

\section{Additional plots of bounds on log-likelihood}
\begin{figure*}[ht]
  \begin{center}
    \begin{tikzpicture}
      \begin{axis}[
          width=0.45*\textwidth,
          height=0.8*0.4\textwidth,
          grid=major, 
          grid style={thick,gray!30},
          xlabel= Steps, 
          ylabel= ELBO,
          legend pos= outer north east,
          x tick label style={ultra thick, anchor=north},
          xmin=10000, xmax=100000,
          ymin=-160, ymax=-100,
          xtick={20000, 40000, 60000, 80000, 100000},
          xticklabels={20K, 40K, 60K, 80K, 100K},
          ytick={-150, -140, -130, -120, -110},
          yticklabels={-150, -140, -130, -120, -110},
          scaled x ticks = false, 
          x tick label style={/pgf/number format/.cd, fixed, fixed zerofill, int detect, 1000 sep={}},
          legend columns=-1,
          legend entries={RT;, R2-G2},
          legend to name={mylegend_2vae_mnist},
          legend style={at={(0.5,-0.2)}, anchor=north,legend columns=2}
        ]
        \addplot [ultra thick, dotted, blue, mark=*,text mark as node=true] 
        table[x=steps,y=elbo,col sep=comma, discard if not={estimator}{RT}]{results/2vae_mnist_elbo_train.csv};
        \addplot [ultra thick, dash pattern=on 6pt off 2pt, orange, mark=+,text mark as node=true] table[x=steps,y=elbo,col sep=comma, discard if not={estimator}{R2-G2}]{results/2vae_mnist_elbo_train.csv};
      \end{axis}
    \end{tikzpicture}
    \begin{tikzpicture}
      \begin{axis}[
          width=0.45*\textwidth,
          height=0.8*0.4\textwidth,
          grid=major, 
          grid style={thick,gray!30},
          xlabel= Steps, 
          ylabel= ELBO,
          legend pos= outer north east,
          x tick label style={ultra thick, anchor=north},
          xmin=10000, xmax=100000,
          ymin=-160, ymax=-100,
          xtick={20000, 40000, 60000, 80000, 100000},
          xticklabels={20K, 40K, 60K, 80K, 100K},
          ytick={-150, -140, -130, -120, -110},
          yticklabels={-150, -140, -130, -120, -110},
          scaled x ticks = false, 
          x tick label style={/pgf/number format/.cd, fixed, fixed zerofill, int detect, 1000 sep={}}
        ]
        \addplot [ultra thick, dotted, blue, mark=*,text mark as node=true] 
        table[x=steps,y=elbo,col sep=comma, discard if not={estimator}{RT}]{results/2vae_mnist_elbo_test.csv};
        \addplot [ultra thick, dash pattern=on 6pt off 2pt, orange, mark=+,text mark as node=true] table[x=steps,y=elbo,col sep=comma, discard if not={estimator}{R2-G2}]{results/2vae_mnist_elbo_test.csv};
      \end{axis}
    \end{tikzpicture}
    \\
    \ref*{mylegend_2vae_mnist}
  \end{center}

  \caption{Bounds on log-likelihood v.s. optimisation steps for a two-layer VAE trained on \emph{MNIST} over 5 runs. We compare the bounds on log-likelihoods when training using the reparameterisation (RT) and R2-G2 estimators on both the training set (left) and test set (right).}
  \label{figure:2vae_mnist_elbo}
\end{figure*}

\begin{figure*}
  \begin{center}
    \begin{tikzpicture}
      \begin{axis}[
          width=0.45*\textwidth,
          height=0.8*0.4\textwidth,
          grid=major, 
          grid style={thick,gray!30},
          xlabel= Steps, 
          ylabel= ELBO,
          legend pos= outer north east,
          x tick label style={ultra thick, anchor=north},
          xmin=10000, xmax=100000,
          ymin=-160, ymax=-125,
          xtick={20000, 40000, 60000, 80000, 100000},
          xticklabels={20K, 40K, 60K, 80K, 100K},
          ytick={-160, -150, -140, -130},
          yticklabels={-160, -150, -140, -130},
          scaled x ticks = false, 
          x tick label style={/pgf/number format/.cd, fixed, fixed zerofill, int detect, 1000 sep={}},
          legend columns=-1,
          legend entries={RT;, R2-G2},
          legend to name={mylegend_2vae_omniglot},
          legend style={at={(0.5,-0.2)}, anchor=north,legend columns=2}
        ]
        \addplot [ultra thick, dotted, blue, mark=*,text mark as node=true] 
        table[x=steps,y=elbo,col sep=comma, discard if not={estimator}{RT}]{results/2vae_omniglot_elbo_train.csv};
        \addplot [ultra thick, dash pattern=on 6pt off 2pt, orange, mark=+,text mark as node=true] table[x=steps,y=elbo,col sep=comma, discard if not={estimator}{R2-G2}]{results/2vae_omniglot_elbo_train.csv};
      \end{axis}
    \end{tikzpicture}
    \begin{tikzpicture}
      \begin{axis}[
          width=0.45*\textwidth,
          height=0.8*0.4\textwidth,
          grid=major, 
          grid style={thick,gray!30},
          xlabel= Steps, 
          ylabel= ELBO,
          legend pos= outer north east,
          x tick label style={ultra thick, anchor=north},
          xmin=10000, xmax=100000,
          ymin=-160, ymax=-125,
          xtick={20000, 40000, 60000, 80000, 100000},
          xticklabels={20K, 40K, 60K, 80K, 100K},
          ytick={-160, -150, -140, -130},
          yticklabels={-160, -150, -140, -130},
          scaled x ticks = false, 
          x tick label style={/pgf/number format/.cd, fixed, fixed zerofill, int detect, 1000 sep={}}
        ]
        \addplot [ultra thick, dotted, blue, mark=*,text mark as node=true] 
        table[x=steps,y=elbo,col sep=comma, discard if not={estimator}{RT}]{results/2vae_omniglot_elbo_test.csv};
        \addplot [ultra thick, dash pattern=on 6pt off 2pt, orange, mark=+,text mark as node=true] table[x=steps,y=elbo,col sep=comma, discard if not={estimator}{R2-G2}]{results/2vae_omniglot_elbo_test.csv};
      \end{axis}
    \end{tikzpicture}
    \\
    \ref*{mylegend_2vae_omniglot}
  \end{center}

  \caption{Bounds on log-likelihood v.s. optimisation steps for a two-layer VAE trained on \emph{Omniglot} over 5 runs. We compare the bounds on  log-likelihoods when training using the reparameterisation (RT) and R2-G2 estimators on both the training set (left) and test set (right).}
  \label{figure:2vae_omniglot_elbo}
\end{figure*}

\begin{figure*}
  \begin{center}
    \begin{tikzpicture}
      \begin{axis}[
          width=0.45*\textwidth,
          height=0.8*0.4\textwidth,
          grid=major, 
          grid style={thick,gray!30},
          xlabel= Steps, 
          ylabel= ELBO,
          legend pos= outer north east,
          x tick label style={ultra thick, anchor=north},
          xmin=10000, xmax=100000,
          ymin=-255, ymax=-235,
          xtick={20000, 40000, 60000, 80000, 100000},
          xticklabels={20K, 40K, 60K, 80K, 100K},
          ytick={-255, -250, -245, -240},
          yticklabels={-255, -250, -245, -240},
          scaled x ticks = false, 
          x tick label style={/pgf/number format/.cd, fixed, fixed zerofill, int detect, 1000 sep={}},
          legend columns=-1,
          legend entries={RT;, R2-G2},
          legend to name={mylegend_2vae_fmnist},
          legend style={at={(0.5,-0.2)}, anchor=north,legend columns=2}
        ]
        \addplot [ultra thick, dotted, blue, mark=*,text mark as node=true] 
        table[x=steps,y=elbo,col sep=comma, discard if not={estimator}{RT}]{results/2vae_fmnist_elbo_train.csv};
        \addplot [ultra thick, dash pattern=on 6pt off 2pt, orange, mark=+,text mark as node=true] table[x=steps,y=elbo,col sep=comma, discard if not={estimator}{R2-G2}]{results/2vae_fmnist_elbo_train.csv};
      \end{axis}
    \end{tikzpicture}
    \begin{tikzpicture}
      \begin{axis}[
          width=0.45*\textwidth,
          height=0.8*0.4\textwidth,
          grid=major, 
          grid style={thick,gray!30},
          xlabel= Steps, 
          ylabel= ELBO,
          legend pos= outer north east,
          x tick label style={ultra thick, anchor=north},
          xmin=10000, xmax=100000,
          ymin=-255, ymax=-235,
          xtick={20000, 40000, 60000, 80000, 100000},
          xticklabels={20K, 40K, 60K, 80K, 100K},
          ytick={-255, -250, -245, -240},
          yticklabels={-255, -250, -245, -240},
          scaled x ticks = false, 
          x tick label style={/pgf/number format/.cd, fixed, fixed zerofill, int detect, 1000 sep={}}
        ]
        \addplot [ultra thick, dotted, blue, mark=*,text mark as node=true] 
        table[x=steps,y=elbo,col sep=comma, discard if not={estimator}{RT}]{results/2vae_fmnist_elbo_test.csv};
        \addplot [ultra thick, dash pattern=on 6pt off 2pt, orange, mark=+,text mark as node=true] table[x=steps,y=elbo,col sep=comma, discard if not={estimator}{R2-G2}]{results/2vae_fmnist_elbo_test.csv};
      \end{axis}
    \end{tikzpicture}
    \\
    \ref*{mylegend_2vae_fmnist}
  \end{center}

  \caption{Bounds on log-likelihood v.s. optimisation steps for a two-layer VAE trained on \emph{Fashion-MNIST} over 5 runs. We compare the bounds on  log-likelihoods when training using the reparameterisation (RT) and R2-G2 estimators on both the training set (left) and test set (right).}
  \label{figure:2vae_fmnist_elbo}
\end{figure*}

\begin{figure*}
  \begin{center}
    \begin{tikzpicture}
      \begin{axis}[
          width=0.45*\textwidth,
          height=0.8*0.4\textwidth,
          grid=major, 
          grid style={thick,gray!30},
          xlabel= Steps, 
          ylabel= ELBO,
          legend pos= outer north east,
          x tick label style={ultra thick, anchor=north},
          xmin=10000, xmax=100000,
          ymin=-165, ymax=-130,
          xtick={20000, 40000, 60000, 80000, 100000},
          xticklabels={20K, 40K, 60K, 80K, 100K},
          ytick={-160, -150, -140},
          yticklabels={-160, -150, -140},
          scaled x ticks = false, 
          x tick label style={/pgf/number format/.cd, fixed, fixed zerofill, int detect, 1000 sep={}},
          legend columns=-1,
          legend entries={RT;, R2-G2},
          legend to name={mylegend_3vae_omniglot},
          legend style={at={(0.5,-0.2)}, anchor=north,legend columns=2}
        ]
        \addplot [ultra thick, dotted, blue, mark=*,text mark as node=true] 
        table[x=steps,y=elbo,col sep=comma, discard if not={estimator}{RT}]{results/3vae_omniglot_elbo_train.csv};
        \addplot [ultra thick, dash pattern=on 6pt off 2pt, orange, mark=+,text mark as node=true] table[x=steps,y=elbo,col sep=comma, discard if not={estimator}{R2-G2}]{results/3vae_omniglot_elbo_train.csv};
      \end{axis}
    \end{tikzpicture}
    \begin{tikzpicture}
      \begin{axis}[
          width=0.45*\textwidth,
          height=0.8*0.4\textwidth,
          grid=major, 
          grid style={thick,gray!30},
          xlabel= Steps, 
          ylabel= ELBO,
          legend pos= outer north east,
          x tick label style={ultra thick, anchor=north},
          xmin=10000, xmax=100000,
          ymin=-165, ymax=-130,
          xtick={20000, 40000, 60000, 80000, 100000},
          xticklabels={20K, 40K, 60K, 80K, 100K},
          ytick={-160, -150, -140},
          yticklabels={-160, -150, -140},
          scaled x ticks = false, 
          x tick label style={/pgf/number format/.cd, fixed, fixed zerofill, int detect, 1000 sep={}}
        ]
        \addplot [ultra thick, dotted, blue, mark=*,text mark as node=true] 
        table[x=steps,y=elbo,col sep=comma, discard if not={estimator}{RT}]{results/3vae_omniglot_elbo_test.csv};
        \addplot [ultra thick, dash pattern=on 6pt off 2pt, orange, mark=+,text mark as node=true] table[x=steps,y=elbo,col sep=comma, discard if not={estimator}{R2-G2}]{results/3vae_omniglot_elbo_test.csv};
      \end{axis}
    \end{tikzpicture}
    \\
    \ref*{mylegend_3vae_omniglot}
  \end{center}

  \caption{Bounds on log-likelihood v.s. optimisation steps for a three-layer VAE trained on \emph{Omniglot} over 5 runs. We compare the bounds on  log-likelihoods when training using the reparameterisation (RT) and R2-G2 estimators on both the training set (left) and test set (right).}
  \label{figure:3vae_omniglot_elbo}
\end{figure*}

\begin{figure*}
  \begin{center}
    \begin{tikzpicture}
      \begin{axis}[
          width=0.45*\textwidth,
          height=0.8*0.4\textwidth,
          grid=major, 
          grid style={thick,gray!30},
          xlabel= Steps, 
          ylabel= ELBO,
          legend pos= outer north east,
          x tick label style={ultra thick, anchor=north},
          xmin=10000, xmax=100000,
          ymin=-260, ymax=-235,
          xtick={20000, 40000, 60000, 80000, 100000},
          xticklabels={20K, 40K, 60K, 80K, 100K},
          ytick={-255, -250, -245, -240},
          yticklabels={-255, -250, -245, -240},
          scaled x ticks = false, 
          x tick label style={/pgf/number format/.cd, fixed, fixed zerofill, int detect, 1000 sep={}},
          legend columns=-1,
          legend entries={RT;, R2-G2},
          legend to name={mylegend_3vae_fmnist},
          legend style={at={(0.5,-0.2)}, anchor=north,legend columns=2}
        ]
        \addplot [ultra thick, dotted, blue, mark=*,text mark as node=true] 
        table[x=steps,y=elbo,col sep=comma, discard if not={estimator}{RT}]{results/3vae_fmnist_elbo_train.csv};
        \addplot [ultra thick, dash pattern=on 6pt off 2pt, orange, mark=+,text mark as node=true] table[x=steps,y=elbo,col sep=comma, discard if not={estimator}{R2-G2}]{results/3vae_fmnist_elbo_train.csv};
      \end{axis}
    \end{tikzpicture}
    \begin{tikzpicture}
      \begin{axis}[
          width=0.45*\textwidth,
          height=0.8*0.4\textwidth,
          grid=major, 
          grid style={thick,gray!30},
          xlabel= Steps, 
          ylabel= ELBO,
          legend pos= outer north east,
          x tick label style={ultra thick, anchor=north},
          xmin=10000, xmax=100000,
          ymin=-260, ymax=-235,
          xtick={20000, 40000, 60000, 80000, 100000},
          xticklabels={20K, 40K, 60K, 80K, 100K},
          ytick={-255, -250, -245, -240},
          yticklabels={-255, -250, -245, -240},
          scaled x ticks = false, 
          x tick label style={/pgf/number format/.cd, fixed, fixed zerofill, int detect, 1000 sep={}}
        ]
        \addplot [ultra thick, dotted, blue, mark=*,text mark as node=true] 
        table[x=steps,y=elbo,col sep=comma, discard if not={estimator}{RT}]{results/3vae_fmnist_elbo_test.csv};
        \addplot [ultra thick, dash pattern=on 6pt off 2pt, orange, mark=+,text mark as node=true] table[x=steps,y=elbo,col sep=comma, discard if not={estimator}{R2-G2}]{results/3vae_fmnist_elbo_test.csv};
      \end{axis}
    \end{tikzpicture}
    \\
    \ref*{mylegend_3vae_fmnist}
  \end{center}

  \caption{Bounds on  log-likelihood v.s. optimisation steps for a three-layer VAE trained on \emph{Fashion-MNIST} over 5 runs. We compare the bounds on  log-likelihoods when training using the reparameterisation (RT) and R2-G2 estimators on both the training set (left) and test set (right).}
  \label{figure:3vae_fmnist_elbo}
\end{figure*}

\section{Licenses}
\label{appendix:licenses}
Codebases:
\begin{itemize}
    \item Convex Potential Flows: Universal Probability Distributions with Optimal Transport and Convex Optimization \citep{huang2021cpflows}: MIT license.
\end{itemize}

Datasets:
\begin{itemize}
    \item MNIST \citep{lecun2010mnist}: Creative Commons Attribution-Share Alike 3.0 license

    \item CIFAR-10 \citep{krizhevsky2009learning}: MIT license

    \item Fashion-MNIST \citep{xiao2017fmnist}: MIT license

    \item Omniglot \citep{lake2015omniglot}: MIT license
\end{itemize}

The conjugate gradient algorithm for each BNN and VAE architecture is modified from the above codebase. All experiments are performed on the above datasets.

\clearpage


\end{document}